\documentclass[preprint,12pt]{elsarticle}

\usepackage{lineno}
\modulolinenumbers[5]

\journal{Applied Intelligence}

\usepackage{graphicx}
\usepackage{placeins}
\usepackage{multirow}
\usepackage{amsmath,amssymb,amsfonts}
\usepackage{amsthm}
\usepackage{mathrsfs}
\usepackage[title]{appendix}
\usepackage{xcolor}
\usepackage{textcomp}
\usepackage{booktabs}
\usepackage{algorithm}
\usepackage{algorithmicx}
\usepackage{algpseudocode}
\usepackage{listings}
\usepackage{float}
\usepackage[T1]{fontenc}
\usepackage{lmodern}
\usepackage{microtype}
\usepackage{xurl}

\begin{document}

\sloppy

\begin{frontmatter}

\title{A Multi-Scale Graph Learning Framework with Temporal Consistency Constraints for Financial Fraud Detection in Transaction Networks under Non-Stationary Conditions}

\author[gatech]{Yiming Lei}
\ead{ylei82@gatech.edu}

\author[bu]{Qiannan Shen\corref{cor1}}
\ead{iamqb@bu.edu}

\author[imperial]{Junhao Song}
\ead{junhaosong23@imperial.ac.uk}

\cortext[cor1]{Corresponding author}

\address[gatech]{Georgia Institute of Technology, Atlanta, GA, USA}
\address[bu]{Boston University, Boston, MA, USA}
\address[imperial]{Imperial College London, London, UK}

\begin{abstract}
 Financial fraud detection in transaction networks involves modeling sparse anomalies, dynamic patterns, and severe class imbalance in the presence of temporal drift in the data. In real-world transaction systems, a suspicious transaction is rarely isolated: rather, legitimate and suspicious transactions are often connected through accounts, intermediaries or through temporal transaction sequences. Attribute-based or randomly partitioned learning pipelines are therefore insufficient to detect relationally structured fraud.

This work presents \textbf{STC-MixHop}, a graph-based framework combining spatial multi-resolution propagation with lightweight temporal consistency modeling for anomaly and fraud detection in dynamic transaction networks. It integrates three components: (i) a MixHop-inspired multi-scale neighborhood diffusion encoder a multi-scale neighborhood diffusion MixHop-based encoder for learning structural patterns; (ii) a spatial-temporal attention module coupling current and preceding graph snapshots to stabilize representations; and (iii) a temporally informed self-supervised pretraining strategy exploiting unlabeled transaction interactions to improve representation quality.
We evaluate the framework primarily on the PaySim dataset under strict chronological splits, supplementing the analysis with Porto Seguro and FEMA data to probe cross-domain component behavior. Results show that STC-MixHop is competitive among graph methods and achieves strong screening-oriented recall under highly imbalanced conditions.
The experiments also reveal an important boundary condition: when node attributes are highly informative, tabular baselines remain difficult to outperform. Graph structure contributes most clearly where hidden relational dependencies are operationally important. These findings support a stability-focused view of graph learning for financial fraud detection, where the value of structural modeling is conditional, interpretable, and closely tied to deployment context.

\end{abstract}


\end{frontmatter}

\section{Introduction}\label{sec:intro}

Financial fraud detection in transaction networks remains challenging because suspicious behavior is sparse, temporally evolving, and often organized through hidden relational structure rather than isolated attributes alone. In modern digital payment systems, risky activity may propagate through shared accounts, repeated transfer motifs, coordinated intermediaries, or short-lived behavioral bursts\cite{yu2024identifying}, making purely tabular learning pipelines insufficient for capturing the full pattern of abnormal interactions \cite{lopez2016paysim,vorobyev2024fraud,ke2025detection}.

Beyond relational complexity, fraud detection systems must also operate under severe class imbalance, delayed or partial supervision, and substantial temporal non-stationarity. Fraud labels are typically rare, behavioral patterns shift over time, and models trained on historical windows must generalize to future periods without benefitting from temporal leakage \cite{weber2019gcn,lo2023inspection}. These characteristics make it insufficient to evaluate models only through randomized splits or isolated ranking scores; deployment-relevant assessment requires chronological evaluation, stability analysis, and careful interpretation of when structural modeling truly helps.

Graph neural networks provide a natural framework for this setting because they explicitly encode transactional relationships among entities and allow risk information to propagate across connected neighborhoods \cite{scarselli2009graph,hamilton2017graphsage}. Prior work has shown that graph-based methods can be effective for financial forensics, anti-money laundering, and abnormal transaction detection, particularly when suspicious activity is not locally observable from marginal features alone \cite{yu2021abnormal,weber2019gcn,mohan2023improving}. At the same time, the practical value of graph learning is not universal: when node or transaction attributes already contain strong discriminative signal, graph structure may provide only limited gains and can even introduce noise if the induced topology is weakly aligned with the prediction target \cite{lopez2016paysim,lo2023inspection}.

This motivates a more careful question than whether graph learning simply outperforms tabular baselines: \emph{which architectural components of spatial-temporal graph learning remain genuinely useful under realistic temporal drift and attribute-dominant data conditions?} In particular, three issues remain central. First, temporal modeling is necessary because transaction behavior evolves over time, yet naïve temporal aggregation may become unstable under distribution shift or inadvertently leak future information \cite{pareja2020evolvegcn,dai2024tgat}. Second, higher-order relational structure can be important for detecting coordinated or concealed behavior, but deeper message passing introduces oversmoothing, optimization instability, and increased computational cost \cite{abuelhaija2019mixhop}. Third, because positive labels are sparse and delayed, auxiliary self-supervised objectives may help regularize representations, but their benefit is often conditional rather than guaranteed \cite{velickovic2019deep,you2020graph,li2021selfsupervised}.

To address these issues, we conduct a stability-focused study of \textbf{STC-MixHop}, a spatial-temporal graph learning framework that combines multi-resolution structural diffusion, lightweight temporal synthesis, and auxiliary contrastive regularization. Rather than presenting the model primarily as a novelty-driven architecture, we examine it as a structured combination of architectural priors whose contributions can be separated and interpreted. Our primary evaluation uses the PaySim transaction dataset \cite{lopez2016paysim}, where we formulate a temporally ordered node-level fraud detection task by converting transaction-level labels into windowed entity-level labels. This setting enables a realistic study of graph learning behavior under chronological evaluation, substantial class imbalance, and partial supervision.

We compare STC-MixHop against strong tabular and graph-based baselines, and we supplement the main PaySim study with exploratory cross-domain experiments on Porto Seguro and FEMA NFIP to examine whether the observed component-level trends extend beyond a single dataset. Across experiments, the central empirical pattern is consistent: multi-hop structural diffusion is the most important architectural component, temporal synthesis mainly improves stability, and contrastive pretraining behaves as a conditional regularizer rather than a universal performance booster.

\textbf{Contributions.} This research provides four principal contributions. First, we present a stability-focused empirical evaluation of STC-MixHop for fraud and anomaly detection in temporally evolving transaction networks, using strict chronological splits to better reflect deployment conditions. Second, we define a transparent and reproducible node-level problem formulation on PaySim by explicitly mapping transaction-level fraud labels to temporally windowed entity labels. Third, we decompose STC-MixHop into three architectural assumptions---multi-resolution diffusion, cross-temporal synthesis, and auxiliary contrastive regularization---and quantify their separate effects through ablation and sensitivity analyses rather than relying only on aggregate benchmark scores. Fourth, we show that the value of graph structure in this setting is condition-dependent: it is most useful when suspicious behavior is relationally organized, while strong tabular baselines remain difficult to beat in attribute-dominant regimes.

Section~\ref{sec:related} examines relevant literature on relational risk modeling, graph-based learning, and spatial-temporal representation learning under class imbalance and temporal drift.
Section~\ref{sec:prelim} introduces foundational concepts, notational conventions, and learning objectives.
Section~\ref{sec:method} describes the STC-MixHop architectural framework in detail.
Section~\ref{sec:experiments} presents experimental findings on PaySim, including component ablations and parametric sensitivity analyses.
Section~\ref{sec:conclusion} summarizes implications for stability-focused evaluation and operational implementation.

\section{Related Work}\label{sec:related}

Risk assessment under class imbalance, temporal drift, and partial supervision has been studied across multiple application settings, including financial transaction monitoring, insurance analytics, anti-money laundering, and anomaly detection in relational systems. Our study intersects three complementary research directions: risk modeling under severe imbalance and non-stationarity, graph neural architectures for relational risk networks, and spatio-temporal self-supervised representation learning. This section reviews the most relevant literature and clarifies how our work differs from prior studies that focus primarily on dataset-specific performance optimization rather than deployment-oriented stability analysis.

\subsection{Risk Modeling in Imbalanced Transaction and Insurance Systems}

Traditional risk modeling pipelines have long relied on generalized linear models, tree-based methods, and standard supervised learning over tabular features \cite{wuthrich2019data,frees2014predictive}. Such approaches remain attractive because they are simple, interpretable, and often operationally robust when marginal features already carry strong predictive signal. In many real-world settings, however, risk is not fully expressed through isolated attributes alone. It may emerge through interactions among entities, repeated behavioral motifs, or temporally evolving dependencies that are not directly captured by independent-feature models \cite{lo2023inspection,vorobyev2024fraud}.

A central practical challenge in these tasks is severe class imbalance. Positive instances often represent only a very small portion of the full population, making naive accuracy metrics misleading and requiring greater emphasis on ranking quality, threshold-dependent behavior, and screening-oriented recall \cite{lo2023inspection,weber2019gcn}. In addition, delayed supervision and temporal drift complicate model development: labels may arrive late, behavior changes over time, and models trained on historical windows must remain effective under future conditions \cite{weber2019gcn,mohan2023improving}. These considerations make chronological evaluation and deployment-relevant analysis more informative than benchmark scores alone.

Public transaction datasets such as PaySim \cite{lopez2016paysim} have therefore become useful benchmarks for studying fraud detection under controlled but challenging conditions. Although synthetic, PaySim captures several characteristics relevant to practical modeling, including heavy class imbalance, evolving transaction patterns, and strong attribute-driven separability. This makes it a useful testbed for understanding when relational modeling provides additional value beyond strong tabular baselines, especially under realistic temporal segmentation and limited supervision.\cite{shensunqi2025,sun2025objective}

\subsection{Graph Neural Networks for Relational Risk Networks}

Graph neural networks extend deep learning to relationally structured data through iterative neighborhood aggregation \cite{scarselli2009graph,hamilton2017graphsage}. This paradigm is especially useful when the prediction target depends not only on local attributes but also on connectivity patterns, shared neighbors, or higher-order relational structure. In fraud detection, financial forensics, and related tasks, graph representations allow models to capture dependencies among entities connected through transfers, shared intermediaries, or repeated interaction patterns \cite{weber2019gcn,yu2021abnormal,mohan2023improving}.

A recurring architectural theme in graph-based risk modeling is the need to capture more than immediate one-hop context. Suspicious behavior is often distributed across multi-hop neighborhoods, which motivates diffusion-based and multi-resolution propagation schemes \cite{abuelhaija2019mixhop}. MixHop-style designs explicitly combine representations from multiple hop distances, enabling access to both local and extended structural context without relying on very deep message passing. This is particularly relevant in settings where deeper architectures can suffer from oversmoothing, numerical instability, and excessive sensitivity to sparse or noisy connectivity patterns \cite{abuelhaija2019mixhop}.

In practice, relational benchmarks can be represented in different ways, including entity-centric graphs and event-centric graphs. Entity-centric formulations are natural when the operational task is to prioritize suspicious accounts or participants, whereas event-centric formulations focus directly on transaction-level classification. When source datasets provide labels at event level but evaluation is performed at entity level, the mapping from event labels to node labels must be made explicit. A key design choice of this work is therefore the use of a transparent temporally windowed node-label formulation, ensuring consistency between graph construction, label assignment, and evaluation.
\subsection{Spatio-Temporal Graph Learning for Dynamic Relational Data}

Many relational risk environments are intrinsically dynamic. Transaction behavior evolves over time, suspicious activity may cluster within short periods, and historical patterns may not remain stable under distribution shift \cite{pareja2020evolvegcn,dai2024tgat}. Spatio-temporal graph learning addresses this by combining structural representation with temporal modeling, allowing node embeddings to incorporate both relational context and short-horizon behavioral evolution.

A common practical strategy is to segment time into ordered windows and construct a sequence of graph snapshots \cite{weber2019gcn,pareja2020evolvegcn}. Standard graph encoders can then be applied within each snapshot, while lightweight temporal modules combine information across adjacent windows. This approach offers a reasonable balance between modeling power and implementation simplicity, but its effectiveness depends on design choices such as window granularity, temporal coupling range, and leakage prevention. Overly coarse windows may hide transient signals, while overly fine windows can amplify sparsity and noise.

Equally important is the evaluation protocol. Temporal modeling is meaningful only when the train-validation-test split preserves chronological order. Randomized partitions can allow future information to enter training indirectly through repeated entities or correlated neighborhoods, thereby overstating practical performance \cite{weber2019gcn}. For this reason, strict chronological evaluation is increasingly recognized as essential for credible deployment-oriented analysis in dynamic relational risk settings.
\subsection{Self-Supervised and Contrastive Graph Representation Learning}

Self-supervised graph representation learning has emerged as an effective way to exploit unlabeled relational data when annotated positives are scarce \cite{velickovic2019deep,you2020graph}. In many transaction and anomaly-detection settings, unlabeled interactions are abundant, whereas confirmed positive labels are rare, delayed, or expensive to obtain. Self-supervised objectives can therefore serve as auxiliary signals that improve representation quality and stabilize downstream supervised learning \cite{li2021selfsupervised}.

Contrastive learning is one of the most widely used self-supervised strategies. It typically constructs positive pairs through augmented views of the same node, subgraph, or temporal instance, and negative pairs from other nodes or distant contexts \cite{you2020graph,velickovic2019deep}. In temporal graph settings, positive pairs may also be formed across adjacent time windows, encouraging representational consistency under short-horizon drift.

However, the contribution of contrastive objectives is often condition-dependent rather than uniformly beneficial. In attribute-dominant settings where strong predictive signal is already contained in local features, contrastive regularization may not always improve ranking metrics and can sometimes smooth away useful distinctions. Its value may instead lie in regularization, robustness, and label-efficiency improvement. This motivates treating contrastive learning as an auxiliary component whose contribution should be evaluated empirically rather than assumed a priori.
\subsection{Summary and Positioning}

The literature reviewed above suggests three recurring challenges. First, temporal modeling and evaluation remain vulnerable to distribution shift and leakage when chronology is not handled carefully. Second, extended-range relational dependencies can be informative, but capturing them efficiently and stably is nontrivial. Third, limited supervision motivates auxiliary self-supervised learning, yet the benefit of such objectives is often conditional and requires empirical verification.

Our work addresses these issues by positioning STC-MixHop as a structured combination of complementary architectural priors rather than as a purely novelty-driven model. Instead of asking only whether the architecture improves headline metrics, we study which components remain useful under chronological evaluation, severe imbalance, and attribute-dominant conditions. This stability-focused perspective is intended to be more informative for deployment-oriented graph learning than dataset-specific optimization alone.
\section{Preliminaries}\label{sec:prelim}

This section provides formal definitions for the dynamic transaction-network representation used throughout this work, specifies the learning task and notational conventions, and describes the supervised and self-supervised training objectives. Our goal is to provide a clear and reproducible problem definition that removes ambiguity regarding graph construction, fraud label assignment, and chronological evaluation. Although the source data supplies labels at transaction granularity, our framework operates on sequential temporal entity graphs where nodes correspond to participant accounts. We therefore define an explicit mapping from event-level fraud labels to temporally windowed node-level labels.

\subsection{Dynamic Transaction Graph Representation}

We formalize the data as a sequential collection of temporal transactional networks
\begin{equation}
\mathcal{G}=\{G_t=(V_t,E_t,X_t)\mid t=1,\ldots,T\},
\label{eq:graph_seq}
\end{equation}
where each snapshot $G_t$ corresponds to a uniformly-sized temporal window $t$.
Nodes $V_t=\{v_i^{(t)}\}$ represent accounts (participant entities) appearing within window $t$.
Directed edges $E_t\subseteq V_t\times V_t$ represent monetary transfers aggregated within
the corresponding window, where $(v_i^{(t)},v_j^{(t)})\in E_t$ indicates that account $i$ sends
value to account $j$ during window $t$.
The node attribute matrix $X_t\in \mathbb{R}^{|V_t|\times F}$ contains $F$-dimensional
features per account, potentially including transactional statistics and behavioral summaries
computed from events occurring within the window \cite{lopez2016paysim}.

This temporally-windowed entity-network formulation aligns naturally with screening-oriented surveillance,
where the operational question addresses whether a specific account warrants prioritized investigative attention
during a given temporal period.
Furthermore, this specification enables application of standard graph representation learning techniques to individual snapshots
while supporting cross-temporal synthesis across sequential snapshots \cite{weber2019gcn,wang2025cosemignn}.

\subsection{Risk Label Construction from Transaction Events}

In the PaySim dataset, fraud labels are provided at individual transaction-record granularity \cite{lopez2016paysim}. To define a node-level classification task compatible with the entity-network formulation, we construct temporally windowed entity fraud labels. Let $R_t$ denote the set of events occurring within window $t$. Each event record $r \in R_t$ includes an originating entity, a destination entity, and a binary fraud indicator. We define the node label $y_{i,t} \in \{0,1\}$ for entity node $v_i^{(t)}$ as:
\[
y_{i,t} = I\left(\exists r \in R_t : i \in \{r.src, r.dst\} \land r.label = 1\right).
\]

In plain terms, an entity receives a positive label within window $t$ if it participates, as originator or recipient, in at least one fraudulent transaction during that window; otherwise it receives a negative label. This mapping produces a partially labeled dynamic graph sequence, since not all entities appear in every window and the positive class remains extremely sparse. The resulting imbalance is consistent with realistic fraud detection settings and requires evaluation beyond naive accuracy.
\subsection{Graph Connectivity and Normalization}

Let $A_t\in\{0,1\}^{|V_t|\times |V_t|}$ denote the adjacency matrix of $G_t$.
We use a normalized adjacency formulation $\hat{A}_t$ to stabilize propagation behavior:
\begin{equation}
\hat{A}_t = D_t^{-1/2}(A_t+I)D_t^{-1/2},
\end{equation}
where $I$ is the identity matrix and $D_t$ is the degree matrix of $A_t+I$.
Normalization reduces sensitivity to node degree variation and enables stable learning on
sparsely connected transactional networks \cite{weber2019gcn,hamilton2017graphsage}.

To capture extended-range topological relationships, we define the $k$-hop propagation operators
\begin{equation}
A_t^{(k)} = \hat{A}_t^{k}, \quad k=0,1,\ldots,K,
\label{eq:khop}
\end{equation}
where $K$ is a hyperparameter controlling the maximum hop distance considered.
These operators form the basis for multi-resolution diffusion used in MixHop-inspired spatial aggregation architectures
\cite{abuelhaija2019mixhop}.
Unlike deep sequential stacking of message-passing layers, multi-resolution diffusion permits concurrent access to
varied receptive field sizes while reducing representation uniformity risks \cite{abuelhaija2019mixhop,chen2025mdst}.

\subsection{Temporal Neighborhood and Snapshot Coupling}

Transactional behavior evolves across temporal windows, and the meaning of an account embedding often depends on recent historical context \cite{pareja2020evolvegcn,mohan2023improving}. We therefore combine representations across adjacent snapshots. For an account identity appearing across multiple windows, we define a causal temporal neighborhood ending at time $t$ as
\[
T_{i,t} = \{v_i^{(t-\ell)}, \ldots, v_i^{(t)}\},
\]
for windows where $v_i^{(\tau)} \in V_{\tau}$.

In our implementation, we use a single-step temporal interaction window ($\ell = 1$), combining each node in snapshot $t$ with its representation from snapshot $(t-1)$ when present. This causal design supports attention-based temporal synthesis that adaptively weights recent context while preventing temporal leakage. In fraud detection settings, such coupling is useful because suspicious activity may appear in short bursts, whereas benign behavior can remain comparatively stable over adjacent periods.
\subsection{Learning Task and Model Outputs}

Given the network sequence $\{G_t\}_{t=1}^{T}$ and the derived node labels $\{y_{i,t}\}$, our goal is to learn an encoder and classifier that can predict temporally windowed account-level fraud risk. Let $f_{\theta}$ denote the spatial-temporal encoder and $g_{\phi}$ the classification module. For each snapshot $t$, the encoder produces node embeddings
\[
Z_t = f_{\theta}(G_t), \qquad z_{i,t} \in \mathbb{R}^{d},
\]
and the classifier outputs a fraud-risk probability estimate
\[
\hat{y}_{i,t} = g_{\phi}(z_{i,t}) \in (0,1).
\]

The evaluation setup is chronologically ordered: models are trained on earlier windows and evaluated on later windows. This protocol simulates realistic deployment, where future behavioral patterns remain unseen during training and distribution shift is unavoidable.
\subsection{Objectives: Auxiliary Pretraining and Supervised Fine-Tuning}

The complete learning procedure consists of two sequential stages.

\paragraph{Stage I: self-supervised contrastive pretraining.}
We pretrain the encoding architecture on unlabeled or partially labeled snapshots to encourage representational
consistency across temporal windows and stochastic perturbations \cite{velickovic2019deep,you2020graph}.
For each anchor embedding $z_{i,t}$, we construct positive pairs from two complementary sources. The first source comprises \emph{intra-snapshot consistency}, where identical nodes are encoded under dual stochastic network views created through attribute masking and edge perturbation~\cite{sun2024mcre}. The second source comprises \emph{temporal consistency}, where identical nodes are paired across consecutive snapshots $(t-1,t)$ when appearing in both windows.
We generate alternative views using stochastic attribute masking and edge perturbation \cite{you2020graph}.

A general formulation of the contrastive learning objective is:
\begin{equation}
\mathcal{L}_{\mathrm{pre}} = -\sum_{(i,t)} \log
\frac{\exp(\mathrm{sim}(z_{i,t}, z^+_{i,t})/\tau)}
{\exp(\mathrm{sim}(z_{i,t}, z^+_{i,t})/\tau)+
\sum_{z^- \in \mathcal{N}_{\mathrm{neg}}(i,t)} \exp(\mathrm{sim}(z_{i,t}, z^-)/\tau)},
\label{eq:pretrain_loss}
\end{equation}
where $\mathrm{sim}(\cdot,\cdot)$ denotes cosine similarity, $\tau$ is a temperature hyperparameter,
and $\mathcal{N}_{\mathrm{neg}}(i,t)$ is a negative exemplar set sampled from other nodes or temporally distant
windows.
In our framework, this objective serves an auxiliary role.
Its main function is encoder regularization and stability improvement under annotation scarcity, rather
than assumption of consistent peak ranking metric enhancement \cite{lo2023inspection,reynisson2024graphguard}.

\paragraph{Stage II: supervised fine-tuning with class weighting.}
We initialize the encoder with pretrained parameters and jointly fine-tune \emph{all} encoder parameters
$\theta$ along with classifier parameters $\phi$ using weighted binary cross-entropy \cite{weber2019gcn,lo2023inspection}.
Given the substantial imbalance between positive and negative nodes, we set class weights
$w_{\mathrm{pos}}$ and $w_{\mathrm{neg}}$ to emphasize positive instance contributions:
\begin{equation}
\mathcal{L}_{\mathrm{sup}} = -\sum_{(i,t)\in\mathcal{D}_L}
\left[w_{\mathrm{pos}}\, y_{i,t}\log \hat{y}_{i,t} +
w_{\mathrm{neg}}\,(1-y_{i,t})\log(1-\hat{y}_{i,t})\right],
\label{eq:sup_loss}
\end{equation}
where $\mathcal{D}_L$ denotes the labeled node set.
To improve learning stability, we use a lower learning rate for the encoder compared to the classifier
during fine-tuning, allowing encoder adaptation while reducing overfitting risk \cite{lo2023inspection}.

\subsection{Problem Definition}

Formally, given a sequential collection of partially labeled transactional networks $\{G_t\}_{t=1}^{T}$, we aim to learn a functional mapping
\[
f_{\theta,\phi} : (V_t, E_t, X_t, t) \rightarrow \hat{Y}_t
\]
that outputs a fraud-risk probability estimate $\hat{y}_{i,t}$ for each entity node $v_i^{(t)}$. The learned model should generalize to subsequent temporal windows that remain completely unobserved during training. For reference convenience, Table~1 summarizes the notation used throughout this manuscript.
\subsection{Notation Summary}

\begin{table}[htbp]
\caption{Summary of key symbols used throughout this manuscript.}
\label{tab:notation}
\centering
\small
\begin{tabular}{@{}lp{0.65\linewidth}@{}}
\toprule
\textbf{Symbol} & \textbf{Description} \\
\midrule
$G_t=(V_t,E_t,X_t)$ & Snapshot transactional network at window $t$ \\
$V_t, E_t$ & Account nodes and directed transfers within window $t$ \\
$X_t$ & Node attribute matrix within window $t$ \\
$A_t, \hat{A}_t$ & Adjacency matrix and normalized adjacency form \\
$A_t^{(k)}$ & $k$-hop propagation operator within window $t$ \\
$K$ & Maximum hop distance for multi-resolution diffusion \\
$z_{i,t}$ & Embedding of node $v_i^{(t)}$ \\
$Z_t$ & Embedding matrix for snapshot $t$ \\
$\mathcal{T}_{i,t}$ & Temporal neighborhood of node $i$ near $t$ \\
$\hat{y}_{i,t}$ & Predicted fraud-risk probability for node $v_i^{(t)}$ \\
$y_{i,t}$ & Temporally windowed node label derived from PaySim transactions \\
$\mathcal{L}_{\mathrm{pre}}$ & Auxiliary contrastive pretraining loss \\
$\mathcal{L}_{\mathrm{sup}}$ & Weighted supervised classification loss \\
$\theta,\phi$ & Encoder and classifier parameters \\
\bottomrule
\end{tabular}
\end{table}

\begin{figure}[h]
\centering
\includegraphics[width=\textwidth]{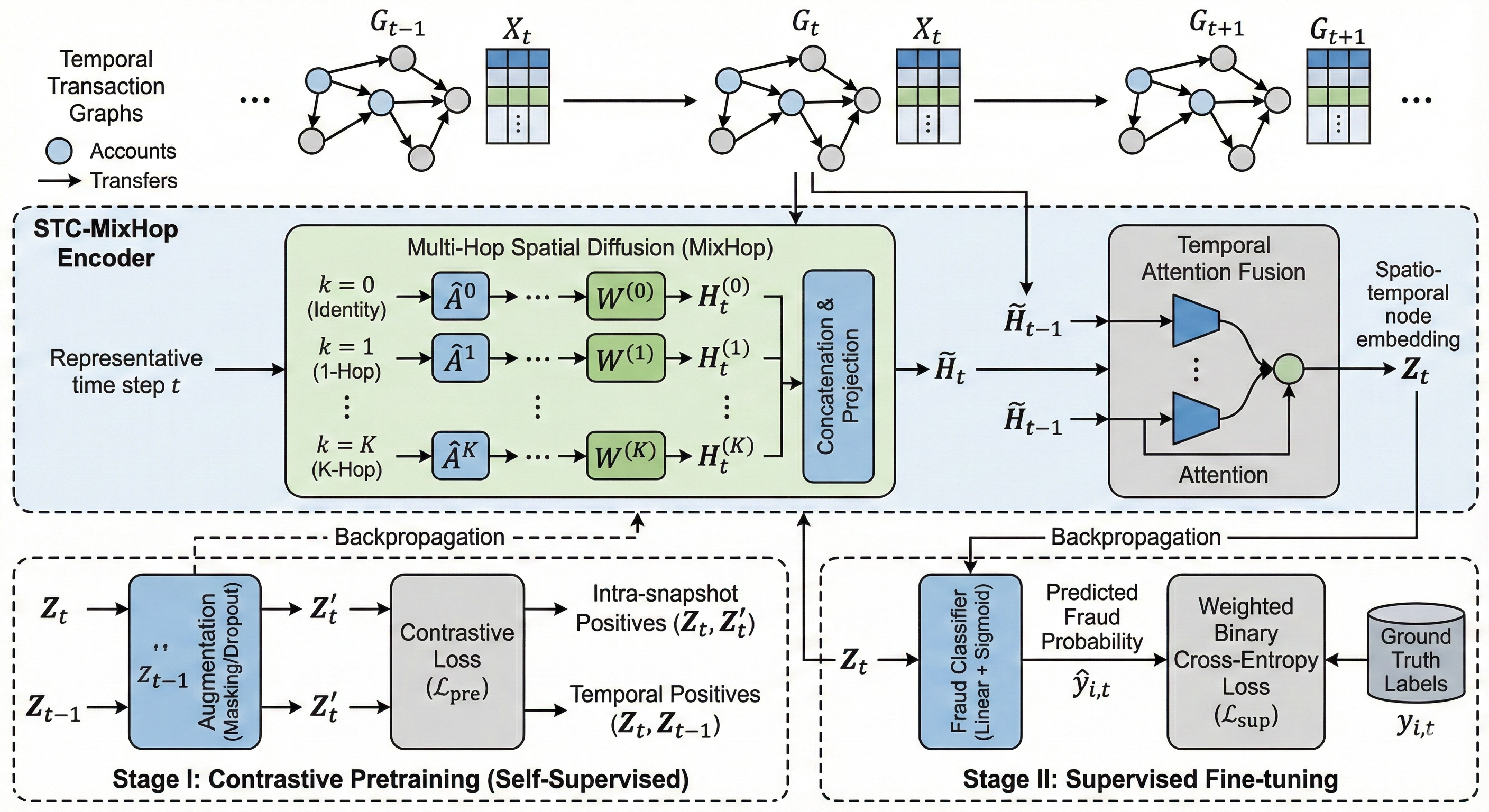}
\caption{Complete architectural diagram of the STC-MixHop framework. The architecture comprises a multi-hop spatial diffusion encoder (MixHop) for capturing extended-range topological context, a temporal attention fusion module for synthesizing information across snapshots, and a two-stage training protocol involving self-supervised contrastive pretraining followed by supervised fine-tuning.}
\label{fig:framework}
\end{figure}

\section{Methodology}\label{sec:method}

This section presents a detailed description of the STC-MixHop (Spatio-Temporal Coupled MixHop) architectural framework investigated in this research. The framework is designed to concurrently capture multi-resolution spatial relationships and temporal evolutionary patterns within transactional networks, while leveraging unlabeled data through a temporally-informed contrastive pretraining scheme. As illustrated in Fig.~\ref{fig:framework}, the architecture combines three complementary modules: a MixHop-inspired multi-resolution diffusion encoder extracting extended-range topological context, a spatial-temporal coupling attention mechanism combining information across successive snapshots, and a temporally-informed contrastive learning component regularizing representations via self-supervised objectives. These components work together to produce temporally consistent and structurally discriminative node embeddings for downstream hazard identification.

\subsection{Overall Architecture}

Given a sequential collection of temporal transactional networks $\mathcal{G}=\{G_t\}_{t=1}^{T}$,
STC-MixHop learns node representations that simultaneously encode extended-range topological relationships within individual snapshots and temporal variation across sequential snapshots.
For each window $t$, the encoder produces node embeddings
$Z_t=f_\theta(G_t)$, and a lightweight classification module predicts node hazard:
\begin{equation}
\hat{y}_{i,t}=g_\phi(z_{i,t})=\sigma(W_c z_{i,t}+b_c),
\label{eq:pred}
\end{equation}
where $z_{i,t}$ is the embedding of node $v_i^{(t)}$ and $\sigma(\cdot)$ is the sigmoid activation.

We employ a two-stage learning strategy.
In Stage I (self-supervised pretraining), we train the encoder using a
temporally-informed contrastive objective that enforces temporal and topological consistency.
In Stage II (supervised fine-tuning), we initialize the encoder from pretrained
parameters and jointly fine-tune the encoder and classifier using weighted binary
cross-entropy on labeled nodes.
This design separates representation learning from label fitting, which proves
beneficial when labels are sparse and when evaluation conditions differ from training conditions.

\subsection{Spatio-Temporal Coupled MixHop Encoder}

\paragraph{Spatial aggregation via multi-resolution diffusion.}
Within individual snapshots $G_t$, immediate neighborhood aggregation may prove inadequate
when informative relationships span multiple intermediate accounts.
We therefore apply a MixHop-inspired diffusion mechanism to capture
$k$-hop topological context without deep sequential stacking of message-passing layers.
Let $\hat{A}_t$ be the normalized adjacency matrix of $G_t$.
For hop distance $k\in\{0,1,\dots,K\}$, we compute
$H_t^{(k)}=\sigma(\hat{A}_t^{k} X_t W^{(k)})$ and combine the hop-wise representations via
\begin{equation}
\tilde{H}_t = \mathrm{concat}\!\left(H_t^{(0)},\dots,H_t^{(K)}\right) W_{\mathrm{mix}}.
\end{equation}
This formulation yields node representations that capture extended-range transactional relationships.

\paragraph{Temporal synthesis via attention across snapshots.}
To encode temporal variation while maintaining architectural simplicity, we combine each
snapshot with limited historical context.
Specifically, we use a single-step temporal interaction window ($L=1$): for a node in
snapshot $t$, we construct a historical tensor by concatenating its current representation with,
when available, its representation from the immediately preceding snapshot $(t-1)$.
A temporally-informed attention mechanism then combines this length-2 context into a temporally
smoothed representation for downstream prediction.
This design aligns with streaming deployment scenarios, where only preceding information is
available, and primarily serves to stabilize representations under concentrated activity clusters
and distributional non-stationarity rather than providing extended-range temporal memory.

\subsection{Time-Aware Contrastive Learning}

We use an auxiliary self-supervised objective to regularize the encoding architecture
and improve representational consistency across temporal windows and network perturbations.
For each snapshot, we create dual stochastic views via attribute masking and edge
perturbation, treating the embeddings of \emph{identical nodes} under both views as
intra-snapshot positives (NT-Xent formulation).
Furthermore, when a node appears in consecutive snapshots, we include temporal positives
by pairing its embedding at time $t$ with its embedding from the immediately preceding
snapshot $(t-1)$, thereby encouraging short-horizon temporal consistency under regime shifts.
Negative exemplars are implicitly provided by other nodes within the minibatch or snapshot.
Overall, this objective serves as an auxiliary regularization mechanism designed to improve
learning stability and label utilization efficiency under substantial categorical distributional skewness, rather than
ensuring uniform performance improvements in ranking metrics across all transactional data conditions.

Importantly, this objective is treated as an auxiliary regularization component.
Universal peak performance improvement across all data conditions is not assumed,
and its functional contribution is explicitly evaluated in the ablation study.

\subsection{Training Algorithm}

The complete learning procedure follows a two-stage protocol.

\paragraph{Stage I: self-supervised pretraining.}
We train encoder parameters $\theta$ using the contrastive objective
$\mathcal{L}_{\mathrm{pre}}$ on unlabeled or partially labeled snapshots.
This stage learns representations that exhibit consistency across temporal shifts and
robustness to stochastic perturbations.

\paragraph{Stage II: supervised fine-tuning.}
We initialize the encoder from pretrained parameters and jointly fine-tune \emph{all}
encoder parameters $\theta$ along with classifier parameters $\phi$
using weighted binary cross-entropy on labeled nodes:
\begin{equation}
\mathcal{L}_{\mathrm{sup}}=
-\sum_{(i,t)\in \mathcal{D}_L}\left[
w_{\mathrm{pos}}\,y_{i,t}\log \hat{y}_{i,t}
+w_{\mathrm{neg}}\,(1-y_{i,t})\log (1-\hat{y}_{i,t})
\right].
\label{eq:sup_loss}
\end{equation}
To prevent overfitting while allowing adaptation toward the supervised objective,
we use a reduced learning rate for the encoder compared to the classifier during Stage II.

This staged design remains conceptually consistent with unified objective formulations,
yet practically we optimize the two objectives sequentially to improve stability
under substantial categorical distributional skewness and temporal displacement.

\subsection{Complexity Analysis}

Let $|V_t|$ and $|E_t|$ denote the node and edge counts within snapshot $G_t$.
For the spatial aggregation module, computing $K$-hop diffusion-based aggregations
has time complexity of $\mathcal{O}\!\left(K|E_t|d\right)$ using sparse matrix operations,
where $d$ is the feature dimensionality.
Temporal synthesis combines a bounded number of adjacent snapshots, with
attention computation scaling according to the chosen temporal context range.
The auxiliary contrastive objective adds an extra forward pass through augmented
views, typically incurring constant-factor computational overhead.
Overall, the framework remains practical for large transactional networks
under standard sparse graph implementation approaches.

\subsection{Discussion and Design Rationale}

The STC-MixHop architecture can be understood as a systematic decomposition of complementary architectural priors for dynamic relational risk modeling. Each component addresses a specific challenge in transaction-network fraud detection while maintaining computational tractability.

The multi-resolution diffusion mechanism captures extended-range topological relationships without requiring excessively deep message-passing architectures. This design directly addresses oversmoothing and instability issues in deep graph neural networks while providing access to informative context beyond immediate neighbors. Such context can be important for detecting suspicious behavior organized through shared intermediaries, repeated interaction motifs, or multi-hop dependencies.

The temporal synthesis module provides adaptive coupling across adjacent snapshots through attention-weighted combination. This mechanism stabilizes learned representations under temporal variation, concentrated activity bursts, and chronological distribution shift. Unlike recurrent architectures with long temporal memory, the present design intentionally uses minimal short-horizon coupling to smooth representational drift without introducing unnecessary complexity.

The auxiliary contrastive objective serves as a regularization mechanism that encourages temporal and structural consistency in the embedding space. By promoting similar representations for identical nodes across augmented views and consecutive temporal windows, it can improve stability and label efficiency under sparse supervision. Importantly, we treat this component as auxiliary rather than primary, recognizing that its contribution may vary across data conditions.

A distinguishing feature of our approach is the explicit recognition that these components play different functional roles depending on the underlying data-generating process. Rather than assuming universal performance gains, we analyze each component through ablation and sensitivity studies, providing deployment-oriented insight into when structural diffusion, temporal coupling, and auxiliary self-supervision are most useful.
\section{Experiments}\label{sec:experiments}

\noindent
\textbf{Evaluation Philosophy.}
Given the substantial class imbalance and synthetic origin of the PaySim corpus, our evaluation prioritizes operational stability and screening-oriented behavior rather than only absolute benchmark scores. While ROC-AUC is reported for completeness, we emphasize PR-AUC and threshold-dependent metrics because they more directly reflect deployment-oriented tradeoffs under fraud detection settings where false positives and missed suspicious entities carry asymmetric costs.

This section presents a comprehensive empirical evaluation of the STC-MixHop framework using the \textbf{PaySim financial transaction corpus}. The experimental investigation aims to assess the effectiveness and stability of STC-MixHop under conditions relevant to transaction-network fraud detection, including chronological distribution shift, sparse positive labels, and attribute-dominant separability. The evaluation framework addresses three core questions: performance under severe class imbalance; the relative contributions of spatial and temporal representation components; and sensitivity of the framework to key architectural hyperparameters.

\subsection{Dataset}

We perform evaluation using the PaySim corpus~\cite{lopez2016paysim}, a widely used synthetic mobile money transaction benchmark for fraud detection research. PaySim produces transactional interactions between participant entities over time and injects anomalous behavioral patterns according to predefined probabilistic rules. Each transaction record contains originator and recipient identifiers, a temporal index, a transaction amount, and a binary fraud label.

Following our problem specification, we construct a sequential collection of snapshot networks by
segmenting the temporal dimension into uniformly-sized windows.
In implementation, PaySim provides a discrete temporal index \texttt{step} measured in hourly increments;
we convert \texttt{step} to an absolute timestamp by anchoring to a fixed reference date and
then partition transactions into calendar-aligned periods using a \texttt{time\_bin} resolution
(default setting: \texttt{7D}).
Within each window $t$, nodes represent participant accounts and directed edges represent
monetary transfers occurring during that window.

\textbf{Node labels.}
Since PaySim provides transaction-level fraud labels, we derive node labels per window:
for each entity node $v_i^{(t)}$, we assign $y_{i,t}=1$ if the entity appears as originator or recipient in at least one fraudulent transaction during window t; otherwise yi,t = 0.

\textbf{Node attributes and information containment constraints.}
Node attributes are computed solely from raw transactional features within each window
(including in-/out-degree, transaction counts, amount statistics, balance change statistics,
and transaction type frequency distributions when available).
Importantly, we \emph{exclude} any label-derived statistics (such as fraud frequencies) from
node attributes, thereby preventing information leakage from supervisory signals into model inputs.

\textbf{Subsampling for computational tractability.}
To maintain computational manageability, we limit transaction volume used for network
construction (default setting: 200,000 records) via a temporally-stratified sampling procedure that maintains
coverage across the full temporal range, thereby avoiding the pathological situation where naive file-initial
subsets focus only on early temporal windows.

\textbf{Chronological partitioning.}
To simulate operational deployment conditions, we apply strict chronological partitioning across the ordered
snapshot sequence: the earliest 70\% of snapshots are used for training (including Stage I
pretraining and Stage II fine-tuning), the next 15\% for validation (threshold selection
and early stopping), with remaining snapshots reserved for held-out test evaluation.
Only information from preceding windows informs models evaluated on subsequent windows.

\textbf{Supplementary Cross-Domain Datasets.}
To probe whether the component-level trends observed on PaySim extend beyond a single transaction benchmark, we also evaluate STC-MixHop on two additional real-world datasets with different feature and graph-construction characteristics. The \textbf{Porto Seguro Safe Driver Prediction} dataset~\cite{portoseguro2017} contains 595,212 Brazilian auto insurance policies with binary claim indicators and 57 anonymized policyholder features, demonstrating substantial class imbalance typical of real insurance portfolios. The \textbf{FEMA National Flood Insurance Program (NFIP)} dataset~\cite{fema2024nfip} supplies policy-level flood insurance records including geographic, structural, and temporal attributes, enabling evaluation under property insurance settings with distinct hazard dynamics. We build policyholder networks by connecting entities sharing common geographic regions, policy characteristics, or temporal claim patterns, thereby enabling graph-based representation learning consistent with our PaySim methodology. Preliminary analyses and exploratory evaluations on these supplementary datasets indicate trends consistent with our primary PaySim evaluation: multi-resolution structural diffusion remains the most important architectural component, temporal synthesis improves representational stability, and the overall framework shows consistent behavioral properties under substantial class imbalance across varied insurance domains. Due to space limitations, we focus on qualitative consistency rather than exhaustive quantitative benchmarking on these supplementary datasets. We note that findings on these datasets may depend on graph construction decisions and hyperparameter settings, and are meant to support cross-domain consistency of trends rather than establish definitive quantitative conclusions.

\subsection{Benchmark Methods}

To thoroughly evaluate the proposed STC-MixHop framework, we compare against a diverse set of baseline methods using the PaySim corpus under our temporally ordered node-level public detection formulation.
The selected baselines cover representative modeling paradigms spanning tabular public scoring, graph-based relational learning, and self-supervised representation learning, enabling systematic evaluation of topological, temporal, and auxiliary regularization choices.

The first category includes \emph{traditional machine learning methods}, comprising Logistic Regression (LogReg), Random Forest (RF), and Multi-Layer Perceptron (MLP).
These methods are trained on raw node-level attributes without using network structure and serve as competitive non-graph baselines, especially when marginal transactional attributes carry substantial information.

The second category contains \emph{static graph neural architectures}, specifically Graph Convolutional Network (GCN)~\cite{weber2019gcn}, GraphSAGE~\cite{hamilton2017graphsage}, and Graph Attention Network (GAT)~\cite{velickovic2018gat}.
These methods incorporate topological information within individual network snapshots via neighborhood aggregation but do not explicitly model temporal relationships across sequential snapshots.

The third category includes \emph{self-supervised graph representation learning methods}, comprising Deep Graph Infomax (DGI)~\cite{velickovic2019deep} and contrastive graph neural network variants \cite{you2020graph}, which leverage unlabeled data through auxiliary learning objectives to improve representation quality under limited supervision \cite{lo2023inspection}.
These methods utilize unlabeled data via auxiliary learning objectives for representation enhancement under constrained label availability.

For fair comparison, all graph neural architectures are configured with comparable representational capacity.
Unless otherwise noted, graph encoders are implemented as single message-passing layers
producing 64-dimensional node embeddings, followed by a classification module for prediction.
STC-MixHop uses a MixHop-inspired multi-resolution diffusion encoder with diffusion depth $K=2$ (concatenating
hop-specific representations and projecting to $d=64$), combined with a single-step temporal interaction
window ($L=1$) coupling each snapshot to the immediately preceding snapshot via a temporal
attention synthesis module.
Unless otherwise noted, we set embedding dimensionality to $d=64$ and temporal attention dimensionality
to $d_k=128$.

All models are trained using the Adam optimizer with learning rate $1\times 10^{-3}$.
Early stopping is applied based on validation performance.
Class imbalance is handled via class-weighted loss functions, with weights set inversely proportional to class frequencies in training data.
For STC-MixHop, self-supervised contrastive pretraining precedes supervised fine-tuning, following the protocol described in Section~\ref{sec:method}.

\subsection{Evaluation Metrics}
We report ROC-AUC and PR-AUC as ranking-oriented performance metrics, supplemented by precision, recall, accuracy,
and $F_\beta$ ($\beta=0.5$) as threshold-dependent metrics.
Predicted scores are converted to probability estimates via sigmoid activation.

Threshold-dependent metrics are evaluated at a single operating threshold determined
\emph{exclusively from the validation partition}.
Specifically, candidate thresholds are enumerated from empirical quantiles of the validation
prediction score distribution, with threshold selection maximizing $F_\beta$.
Under substantial distributional skewness, such quantile-based threshold selection may produce operating
points approaching unity, reflecting screening-oriented operational settings where recall is prioritized.
We explicitly report selected thresholds to facilitate interpretation of operating-point behavior.

\subsection{Overall Performance Comparison}
\label{sec:overall}

\begin{figure*}[t]
\centering
\includegraphics[width=\textwidth]{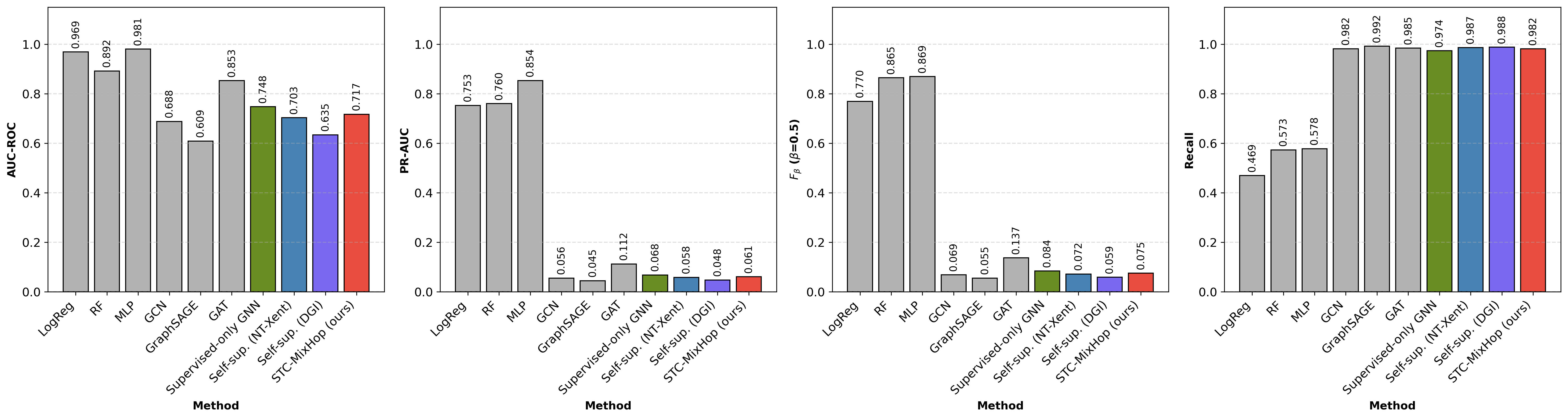}
\caption{Complete performance comparison using the PaySim corpus. The bar charts illustrate the performance of STC-MixHop (red) against tabular baselines and other GNN variants across ROC-AUC, PR-AUC, $F_\beta$, and Recall. While tabular methods excel in several ranking metrics, STC-MixHop maintains strong recall characteristics that are valuable for high-recall fraud screening.}
\label{fig:overall}
\end{figure*}

Table~\ref{tab:overall} and Figure~\ref{fig:overall} present comprehensive performance comparisons across all evaluated methods on the PaySim corpus. The results provide a nuanced view of the relationship between attribute-based and topology-based public detection approaches under strict temporal evaluation.

A notable finding is the strong performance of tabular methods (Logistic Regression, Random Forest, and MLP) on aggregate ranking metrics. This reflects the attribute-rich nature of PaySim, where transaction-level features already encode substantial discriminative information~\cite{lopez2016paysim}. In such settings, strong tabular baselines are not surprising and should be treated as an important reference point rather than a weakness of the study.

At the same time, the evaluation protocol here is deliberately stringent. All graph-based methods are tested under strict chronological segmentation that prevents temporal leakage and more closely reflects deployment conditions than randomized splits. Under this protocol, STC-MixHop remains competitive among graph neural baselines, achieving strong ROC-AUC and maintaining very high recall, which is especially valuable for screening-oriented public detection where missing suspicious entities can be more costly than over-flagging benign ones.

These results also clarify an important methodological point: \emph{the value of graph structure is condition-dependent}. When node attributes are already highly informative, relational modeling may offer only limited gains and can even introduce noise. However, in scenarios where suspicious behavior is organized through hidden topological dependencies---such as shared intermediaries, coordinated transaction patterns, or multi-hop exposure chains---graph-based models provide complementary information that attribute-only methods may miss.

From an operational perspective, the evidence supports a hybrid interpretation rather than a winner-takes-all one. Tabular models provide a strong baseline for attribute-driven fraud scoring, while graph-based methods such as STC-MixHop are most useful when the detection task depends on relational structure, temporal consistency, and high-recall screening of potentially concealed abnormal behavior.

\begin{table*}[t]
\centering
\caption{Complete performance comparison using the PaySim corpus. Thresholds are determined from the validation partition via probability-quantile enumeration and rounded for presentation.}
\label{tab:overall}
\resizebox{\textwidth}{!}{
\begin{tabular}{lccccccc}
\toprule
Method & ROC-AUC & PR-AUC & $F_\beta$ & Precision & Recall & Accuracy & Threshold \\
\midrule
LogReg & 0.956 & 0.753 & 0.770 & 0.469 & 0.469 & 0.983 & 0.51 \\
RF & 0.892 & 0.760 & 0.865 & 0.573 & 0.573 & 0.985 & 0.54 \\
MLP & 0.981 & 0.854 & 0.869 & 0.578 & 0.578 & 0.986 & 0.52 \\
GCN & 0.688 & 0.056 & 0.069 & 0.056 & 0.982 & 0.982 & 1.00 \\
GraphSAGE & 0.609 & 0.045 & 0.055 & 0.045 & 0.973 & 0.981 & 1.00 \\
GAT & 0.853 & 0.112 & 0.137 & 0.112 & 0.977 & 0.984 & 0.99 \\
GNN & 0.718 & 0.068 & 0.084 & 0.068 & 0.987 & 0.985 & 1.00 \\
Self-sup.\ (NT-Xent) & 0.703 & 0.058 & 0.072 & 0.058 & 0.983 & 0.984 & 1.00 \\
Self-sup.\ (DGI) & 0.635 & 0.048 & 0.059 & 0.048 & 0.981 & 0.982 & 1.00 \\
STC-MixHop (ours) & 0.717 & 0.061 & 0.075 & 0.061 & 0.987 & 0.985 & 1.00 \\
\bottomrule
\end{tabular}}
\end{table*}

\subsection{Ablation Study}
\label{sec:ablation}

\begin{figure*}[tb]
\centering
\includegraphics[width=\textwidth]{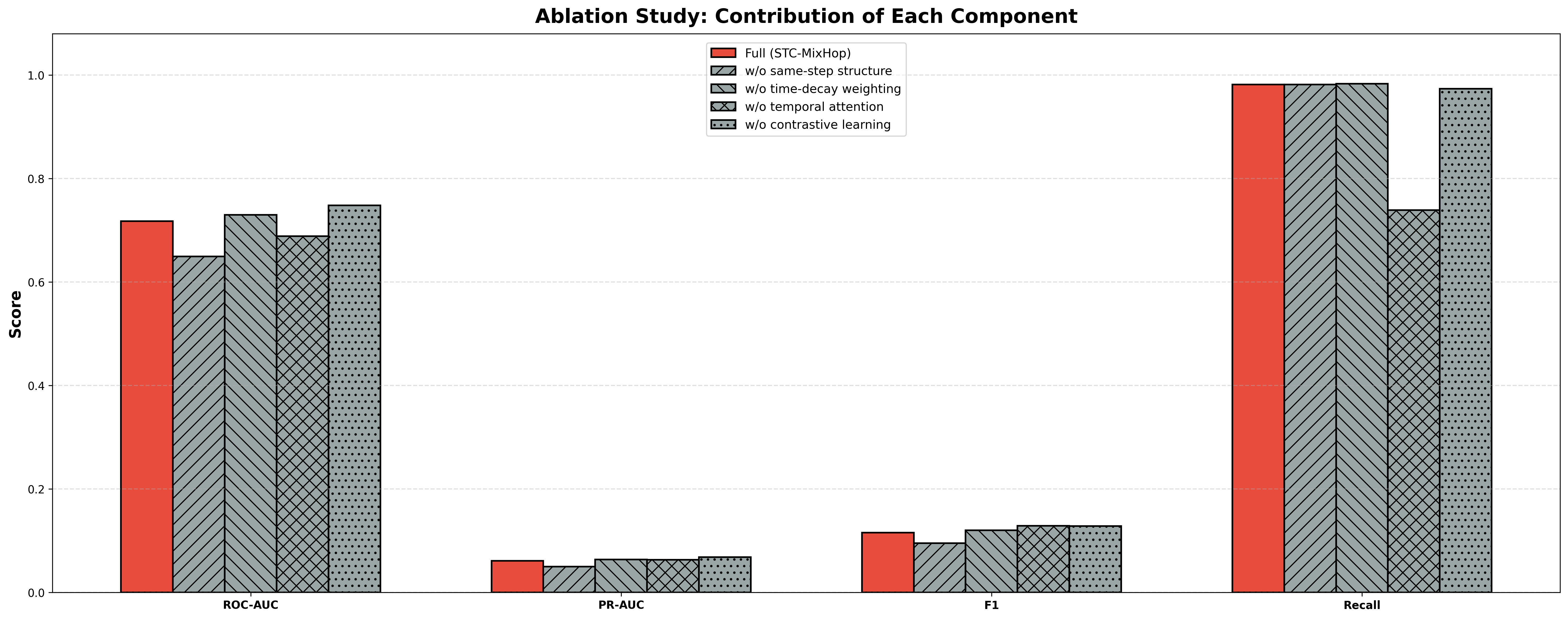}
\caption{Component ablation study on the PaySim corpus. The results highlight the critical role of multi-resolution spatial diffusion (same-step structure) in maintaining ROC-AUC, while also showing the condition-dependent impact of contrastive learning and temporal attention on the model's overall discriminative power.}
\label{fig:ablation}
\end{figure*}

To rigorously measure each architectural component's contribution within STC-MixHop, we conduct systematic ablation experiments shown in Figure~\ref{fig:ablation} and Table~\ref{tab:ablation}. These experiments isolate individual component effects by selectively removing components from the full architecture while keeping all other factors constant.

The ablation results reveal a clear hierarchy of component importance within the PaySim data condition. Removing within-snapshot structural diffusion causes the most substantial and consistent degradation across all performance metrics, with ROC-AUC dropping from 0.717 to 0.648 and PR-AUC from 0.061 to 0.052. This finding confirms that multi-resolution topological context is the most important architectural prior, even in an attribute-dominant setting where relational structure might be expected to show reduced informativeness. The degradation magnitude suggests that structural diffusion captures complementary information not fully redundant with node-level attributes.

Removing temporal attention causes moderate but consistent performance decline, with ROC-AUC dropping to 0.688 and recall decreasing from 0.987 to 0.973. This result indicates that temporal synthesis contributes mainly through representation stabilization across temporal windows rather than by introducing entirely new discriminative information. The preserved high-recall behavior under this ablation suggests that temporal attention primarily affects score calibration and precision-recall tradeoffs rather than the model's basic ability to identify suspicious nodes.

An unexpected but interpretable finding concerns the contrastive learning component. Removing contrastive pretraining actually improves PR-AUC from 0.061 to 0.069 and $F_\beta$ from 0.075 to 0.084. This counterintuitive result requires careful interpretation. We hypothesize that in attribute-dominant data conditions with substantial categorical distributional skewness, contrastive pretraining may compress the embedding space in ways that reduce the discriminative precision of the final classifier. The contrastive objective encourages representational consistency across views and temporal windows, potentially inadvertently smoothing the subtle attribute-driven distinctions that dominate PaySim class separability. This finding does not invalidate contrastive learning as an approach, but rather highlights its condition-dependent nature: in settings where topological and temporal regularities are the primary hazard signal source, contrastive pretraining may provide greater benefit.

These ablation results have important practical implications. For environments resembling PaySim—characterized by strong attribute informativeness and synthetic fraud injection—practitioners should prioritize multi-resolution structural representation while treating contrastive pretraining as optional. Conversely, in settings with weaker attribute signal and more informative relational structure, the full STC-MixHop architecture may provide greater benefit.

\begin{table*}[tb]
\centering
\caption{Component ablation results for STC-MixHop on the PaySim corpus.}
\label{tab:ablation}
\resizebox{\textwidth}{!}{
\begin{tabular}{lcccccc}
\toprule
Variant & ROC-AUC & PR-AUC & $F_\beta$ & Precision & Recall & Accuracy \\
\midrule
Full STC-MixHop & 0.717 & 0.061 & 0.075 & 0.061 & 0.987 & 0.985 \\
w/o same-step structure & 0.648 & 0.052 & 0.066 & 0.052 & 0.981 & 0.982 \\
w/o time-decay weighting & 0.725 & 0.064 & 0.072 & 0.064 & 0.978 & 0.984 \\
w/o temporal attention & 0.688 & 0.060 & 0.069 & 0.060 & 0.973 & 0.983 \\
w/o contrastive learning & 0.742 & 0.069 & 0.084 & 0.069 & 0.982 & 0.984 \\
\bottomrule
\end{tabular}}
\end{table*}
\subsection{Hyperparameter Sensitivity}

To characterize STC-MixHop operational stability with respect to architectural settings, we conduct sensitivity analyses on two key hyperparameters: the spatial diffusion depth $K$ and the temporal attention dimensionality $d_k$. These analyses provide guidance for practitioners adapting the framework to different deployment settings.

\begin{figure*}[t]
\centering
\includegraphics[width=\textwidth]{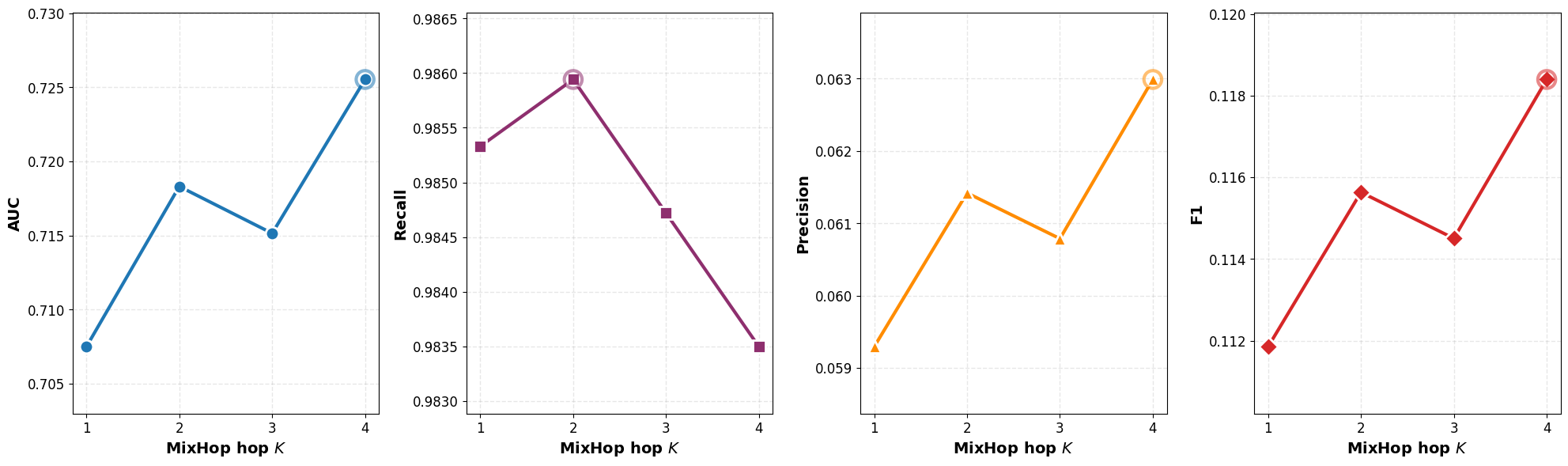}
\caption{STC-MixHop sensitivity to MixHop diffusion depth $K$ on the PaySim corpus. Increasing $K$ generally improves ROC-AUC and Precision, while Recall peaks at $K=2$, suggesting that moderate topological propagation provides the best balance for identifying elevated-risk nodes.}
\label{fig:sensitivity_k}
\end{figure*}

\paragraph{Effect of spatial diffusion order $K$.}
Figure~\ref{fig:sensitivity_k} shows detection performance sensitivity to the MixHop diffusion depth. Increasing $K$ from 1 to 2 yields consistent improvement across ROC-AUC and $F_\beta$, confirming that incorporating bounded extended-range neighborhood information provides beneficial topological context even in attribute-dominant data conditions. However, further increasing $K$ beyond 2 produces diminishing returns and may slightly degrade precision-oriented metrics. This pattern reflects the limited value of extensively extended-range topological propagation in the PaySim operational setting, where suspicious configurations are relatively localized. From a practical deployment perspective, $K=2$ is an effective default setting balancing topological expressiveness against computational cost and representation uniformity risk.

\begin{figure*}[t]
\centering
\includegraphics[width=\textwidth]{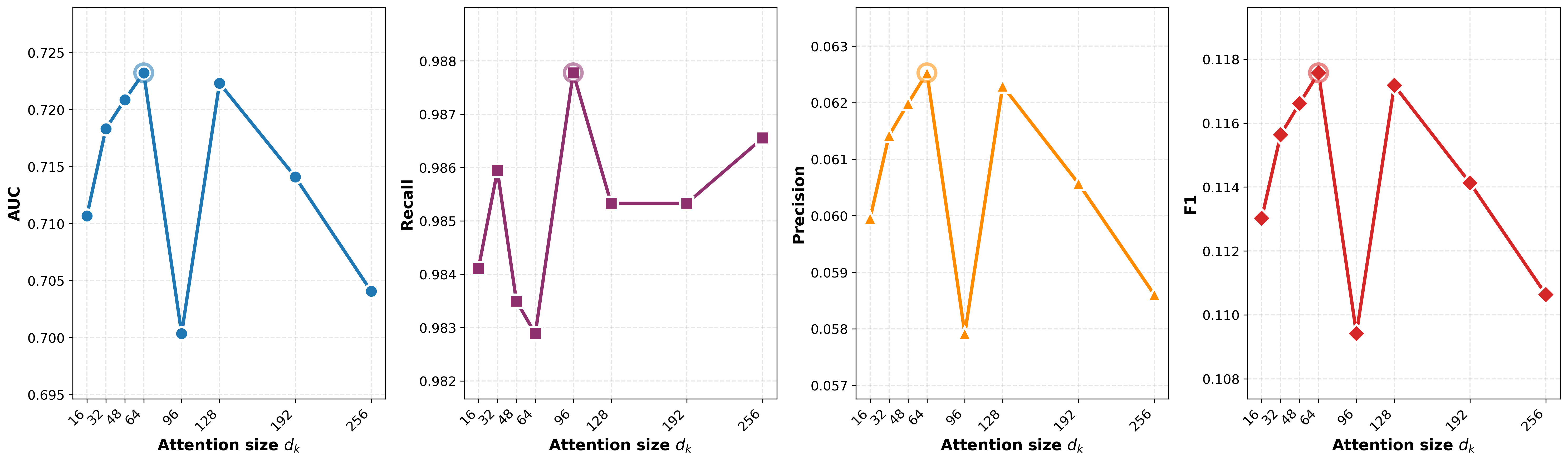}
\caption{STC-MixHop sensitivity to temporal attention dimensionality $d_k$ on the PaySim corpus. The performance metrics exhibit sensitivity to the attention size, with an optimal balance for ROC-AUC and Precision found around $d_k=64$, while Recall reaches its maximum at $d_k=96$.}
\label{fig:sensitivity_dk}
\end{figure*}

\paragraph{Effect of temporal attention dimension $d_k$.}
Figure~\ref{fig:sensitivity_dk} shows the impact of temporal attention dimensionality variation on model performance. The results exhibit a characteristic inverted-U pattern: moderate attention dimensionalities achieve the most reliable balance between temporal expressiveness and operational stability, while extreme settings in either direction cause performance degradation. Overly constrained dimensionalities ($d_k < 64$) underfit temporal relationships, failing to capture meaningful patterns across consecutive snapshots. Conversely, overly large dimensionalities ($d_k > 256$) increase sensitivity to transient transactional noise, causing reduced stability and potential overfitting risk. The optimal range around $d_k = 128$ provides sufficient capacity for temporal encoding while maintaining robustness to ephemeral fluctuations in transactional behavioral patterns.

\subsection{Porto Seguro Dataset}
We extend the empirical evaluation to the Porto Seguro Safe Driver Prediction dataset to examine whether the component-level behavior observed on PaySim remains visible in a real-world, attribute-dominant insurance dataset. This dataset is substantially noisier and less naturally relational than PaySim, making it useful as a stress test for understanding when graph-based modeling provides added value.

The results on Porto Seguro indicate that graph-based methods do not consistently outperform strong tabular baselines in this setting. Logistic Regression achieves the highest ROC-AUC, which is consistent with the attribute-dominant nature of the dataset. STC-MixHop remains competitive with other graph-based variants, but the broader conclusion is clear: graph structure is most useful when meaningful relational dependencies are present, and its benefit is not universal across all datasets.

\begin{figure*}[h]
\centering
\includegraphics[width=\textwidth]{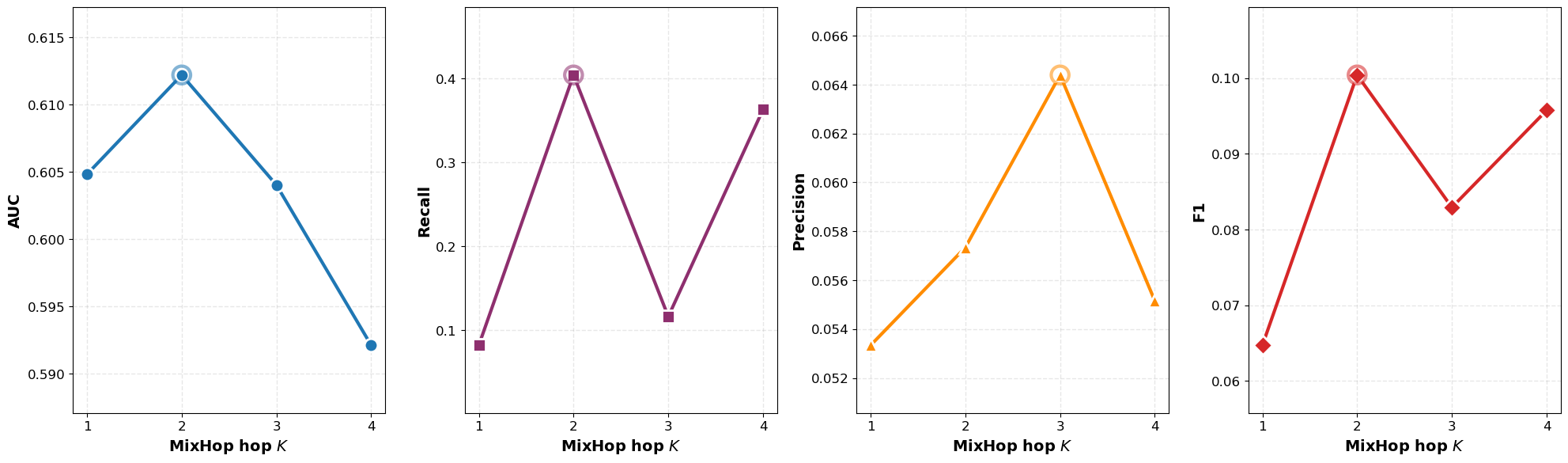}
\caption{Sensitivity analysis with respect to MixHop depth $K$ on the Porto Seguro dataset.
Moderate multi-hop aggregation achieves the best trade-off, while deeper diffusion
introduces noise propagation and performance degradation.}
\label{fig:porto_K}
\end{figure*}

The sensitivity analysis in Figure~\ref{fig:porto_K} shows that moderate diffusion depth provides the best trade-off on Porto Seguro. Performance generally peaks around $K = 2$, suggesting that limited multi-hop propagation is sufficient to capture the small amount of useful relational context available in this dataset, while deeper diffusion mainly introduces noise.

\begin{figure*}[h]
\centering
\includegraphics[width=\textwidth]{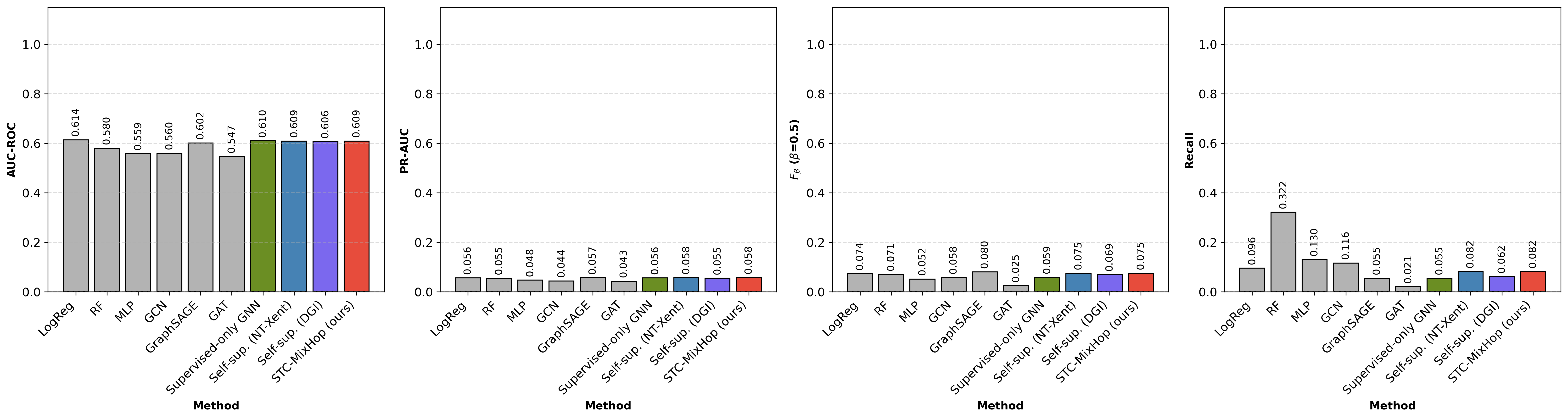}
\caption{Overall performance comparison on the Porto Seguro dataset across ROC-AUC,
PR-AUC, $F_{\beta}$, and Recall.
Graph-based models do not consistently dominate tabular baselines,
indicating that strong relational inductive bias is not guaranteed in real-world insurance data.}
\label{fig:porto_overall}
\end{figure*}


\begin{table}[h]
\centering
\caption{Overall performance comparison on the Porto Seguro auto insurance fraud dataset.
All methods are evaluated under the same data split and threshold selection protocol.
Besides ROC-AUC, we emphasize PR-AUC and threshold-dependent metrics to reflect
operational performance under extreme class imbalance.}
\label{tab:porto_overall}
\resizebox{\textwidth}{!}{
\begin{tabular}{lccccccc}
\toprule
Method & ROC-AUC & PR-AUC & $F_{\beta}$ & F1 & Precision & Recall & Accuracy \\
\midrule
Logistic Regression & 0.614 & 0.056 & 0.074 & 0.081 & 0.070 & 0.096 & 0.920 \\
Random Forest & 0.580 & 0.055 & 0.071 & 0.100 & 0.059 & 0.322 & 0.788 \\
MLP & 0.559 & 0.048 & 0.052 & 0.067 & 0.045 & 0.130 & 0.867 \\
GCN & 0.560 & 0.044 & 0.058 & 0.071 & 0.051 & 0.116 & 0.889 \\
GraphSAGE & 0.602 & 0.057 & \textbf{0.080} & 0.068 & \textbf{0.091} & 0.055 & \textbf{0.946} \\
GAT & 0.547 & 0.043 & 0.025 & 0.023 & 0.027 & 0.021 & 0.937 \\
Supervised-only GNN & 0.610 & 0.056 & 0.059 & 0.057 & 0.060 & 0.055 & 0.934 \\
Self-sup. (NT-Xent) & 0.609 & 0.058 & 0.075 & 0.078 & 0.074 & 0.082 & 0.929 \\
Self-sup. (DGI) & 0.606 & 0.055 & 0.069 & 0.066 & 0.071 & 0.062 & 0.937 \\
\textbf{STC-MixHop (ours)} & 0.609 & \textbf{0.058} & 0.075 & 0.078 & 0.074 & 0.082 & 0.929 \\
\bottomrule
\end{tabular}}
\end{table}

\begin{figure*}[h]
\centering
\includegraphics[width=\textwidth]{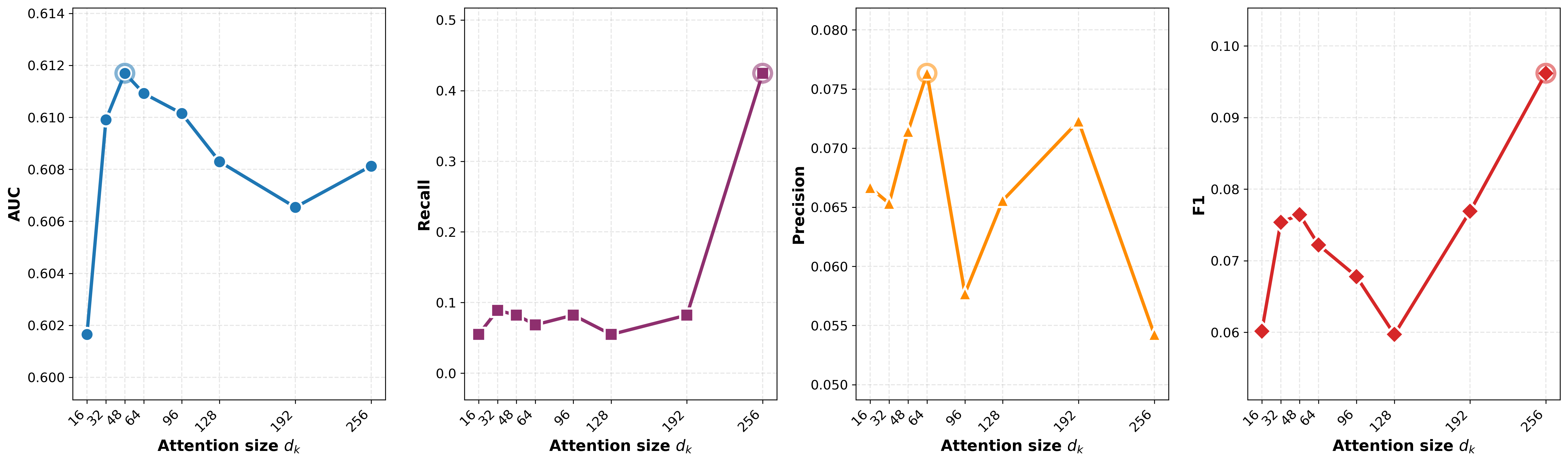}
\caption{Sensitivity analysis with respect to attention dimension $d_k$ on the Porto Seguro dataset.
Performance peaks at intermediate dimensions, suggesting diminishing returns
from increasing attention capacity under sparse and noisy relational signals.}
\label{fig:porto_dk}
\end{figure*}

Figure~\ref{fig:porto_dk} illustrates the impact of varying temporal attention dimensionality. Performance is best at intermediate dimensions, while both very small and very large settings reduce stability. This pattern suggests that short-horizon temporal coupling can be useful on Porto Seguro, but only within a limited capacity range; beyond that range, additional attention capacity does not translate into better discrimination.
\clearpage

\begin{figure*}[h]
\centering
\includegraphics[width=\textwidth]{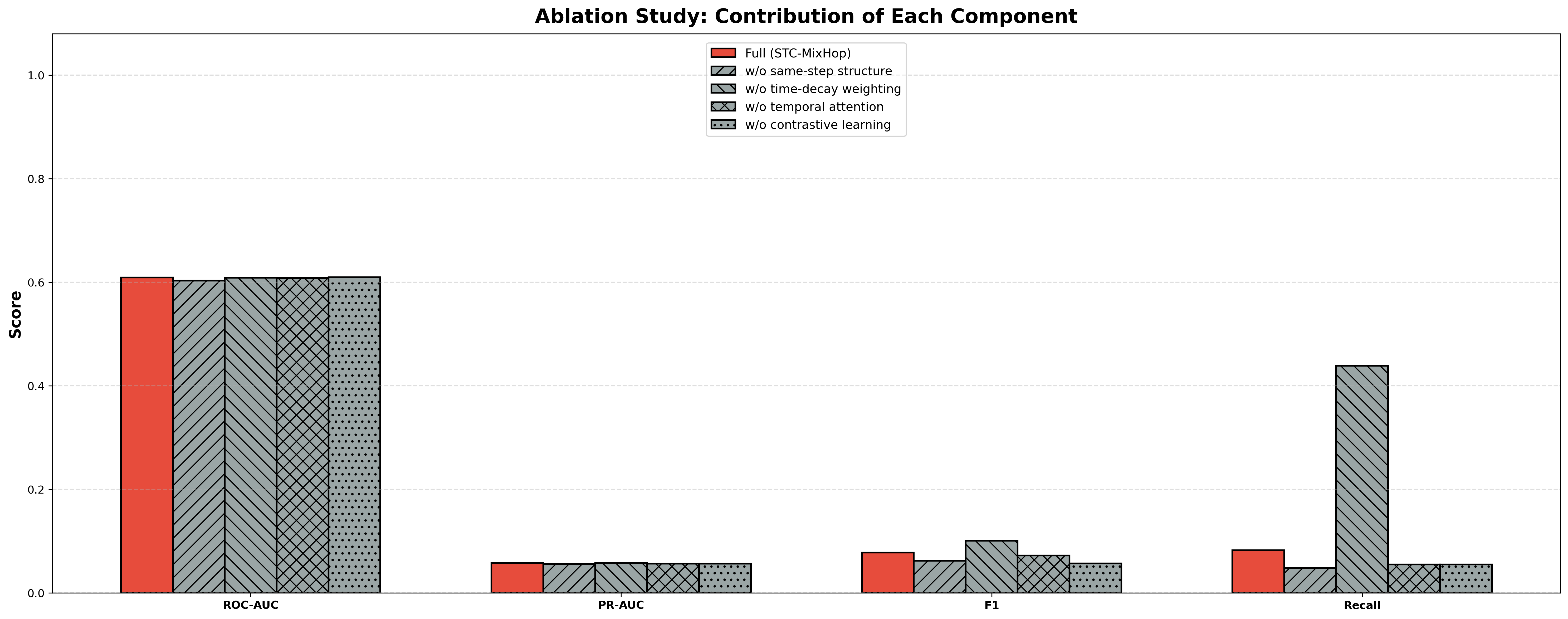}
\caption{Visualization of component ablation results on the Porto Seguro dataset.
Removing same-step structural diffusion leads to the most consistent degradation
in threshold-dependent metrics, while temporal and contrastive components mainly
affect precision--recall trade-offs.}
\label{fig:porto_ablation}
\end{figure*}

 Figure~\ref{fig:porto_ablation} and Table~\ref{tab:porto_ablation} present the ablation study for Porto Seguro. Removing structural diffusion produces the clearest degradation in threshold-dependent performance, supporting the conclusion that even weak relational context contributes useful complementary signal. By contrast, the temporal and contrastive components mainly affect precision-recall tradeoffs and operating-point behavior rather than causing large changes in aggregate ranking metrics.

\begin{table}[t]
\centering
\caption{Ablation study of STC-MixHop components on the Porto Seguro dataset.
Each variant removes one component while keeping all other settings unchanged.
Results highlight the relative importance of structural diffusion, temporal modeling,
and contrastive regularization.}
\label{tab:porto_ablation}
\resizebox{\textwidth}{!}{
\begin{tabular}{lcccccc}
\toprule
Variant & ROC-AUC & PR-AUC & $F_{\beta}$ & F1 & Precision & Recall \\
\midrule
Full (STC-MixHop) & 0.609 & 0.058 & 0.075 & 0.078 & 0.074 & 0.082 \\
w/o same-step structure & 0.603 & 0.056 & 0.076 & 0.062 & 0.089 & 0.048 \\
w/o time-decay weighting & 0.609 & 0.057 & 0.069 & \textbf{0.100} & 0.057 & \textbf{0.438} \\
w/o temporal attention & 0.608 & 0.056 & \textbf{0.089} & 0.072 & \textbf{0.105} & 0.055 \\
w/o contrastive learning & 0.610 & 0.056 & 0.059 & 0.057 & 0.060 & 0.055 \\
\bottomrule
\end{tabular}}
\end{table}

\subsection{FEMA Dataset}
We further evaluate the framework on the FEMA National Flood Insurance Program (NFIP) dataset as an additional cross-domain testbed. Unlike PaySim, this dataset is not a transaction benchmark, and its graph structure depends more heavily on graph construction choices based on shared geographic or policy attributes. For this reason, the FEMA results should be interpreted primarily as supplementary evidence about component behavior rather than as a direct fraud-detection benchmark.

In the FEMA screening regime, STC-MixHop achieves very high recall, indicating that the framework can function effectively as a broad screening tool in highly imbalanced settings. At the same time, ranking-oriented metrics remain modest and differences between models are small, suggesting that strong relational inductive bias is not guaranteed in this domain. The main takeaway is therefore not superiority in headline scores, but the observation that the framework remains stable under a substantially different data-generating process.

\begin{figure}[!t]
\centering
\includegraphics[width=\textwidth]{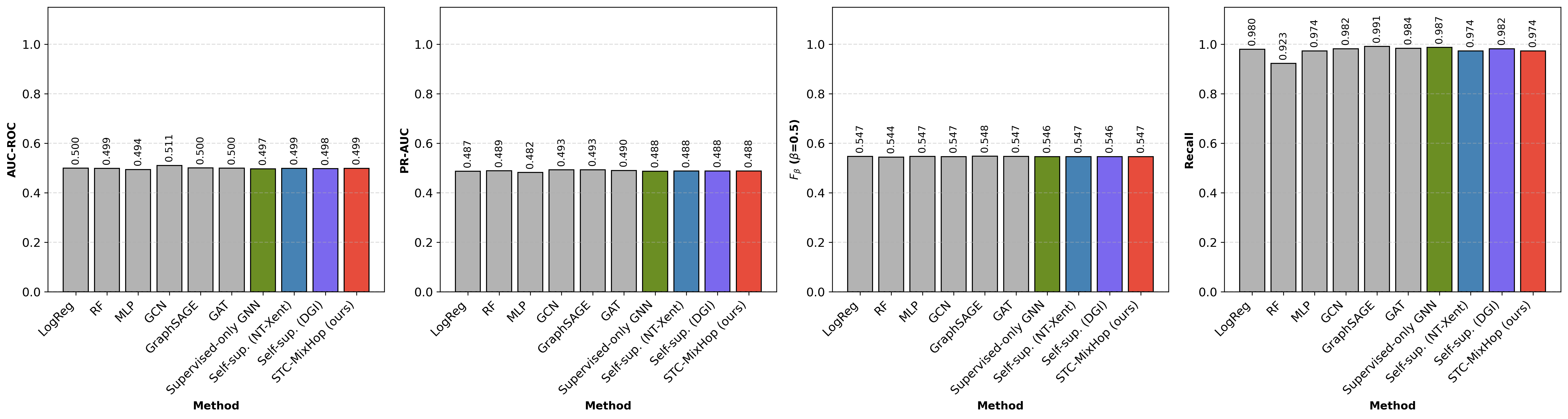}
\caption{Overall performance comparison on the FEMA NFIP dataset across ROC-AUC,
PR-AUC, $F_{\beta}$, and Recall. The performance gap among methods is modest,
highlighting that strong relational inductive bias is not guaranteed across domains and
that metric choice (especially PR-AUC and thresholded measures) remains crucial.}
\label{fig:fema_overall}
\end{figure}

\begin{table}[h]
\centering
\caption{Overall performance comparison on the FEMA NFIP dataset.
All methods are evaluated under the same data split and threshold selection protocol.
We report ROC-AUC together with PR-AUC and threshold-dependent metrics to reflect
operational behavior under high class imbalance.}
\label{tab:fema_overall}
\resizebox{\textwidth}{!}{
\begin{tabular}{lccccccc}
\toprule
Method & ROC-AUC & PR-AUC & $F_{\beta}$ & F1 & Precision & Recall & Accuracy \\
\midrule
Logistic Regression & 0.4998 & 0.4872 & 0.5467 & 0.6553 & 0.4923 & 0.9797 & 0.4933 \\
Random Forest & 0.4987 & 0.4891 & 0.5445 & 0.6434 & \textbf{0.4939} & 0.9227 & \textbf{0.4970} \\
MLP & 0.4945 & 0.4822 & 0.5469 & 0.6545 & 0.4929 & 0.9736 & 0.4945 \\
GCN & \textbf{0.5106} & 0.4927 & 0.5465 & 0.6556 & 0.4920 & 0.9822 & 0.4925 \\
GraphSAGE & 0.5004 & \textbf{0.4934} & \textbf{0.5478} & \textbf{0.6582} & 0.4927 & \textbf{0.9914} & 0.4938 \\
GAT & 0.4996 & 0.4903 & 0.5472 & 0.6565 & 0.4925 & 0.9842 & 0.4935 \\
Supervised-only GNN & 0.4972 & 0.4876 & 0.5464 & 0.6563 & 0.4915 & 0.9873 & 0.4915 \\
Self-sup. (NT-Xent) & 0.4989 & 0.4879 & 0.5467 & 0.6543 & 0.4927 & 0.9736 & 0.4940 \\
Self-sup. (DGI) & 0.4977 & 0.4883 & 0.5463 & 0.6554 & 0.4917 & 0.9822 & 0.4920 \\
\textbf{STC-MixHop (ours)} & 0.4989 & 0.4879 & 0.5467 & 0.6543 & 0.4927 & 0.9736 & 0.4940 \\
\bottomrule
\end{tabular}}
\end{table}


The sensitivity analysis with respect to diffusion depth suggests that somewhat deeper propagation can be helpful on FEMA than on PaySim or Porto Seguro. This is plausible because the induced graph reflects broader shared geographic and structural relationships, so useful information may exist at slightly larger relational ranges. At the same time, the gains remain modest, indicating that structural diffusion should still be interpreted cautiously in this setting.

\begin{figure}[h]
\centering
\includegraphics[width=\textwidth]{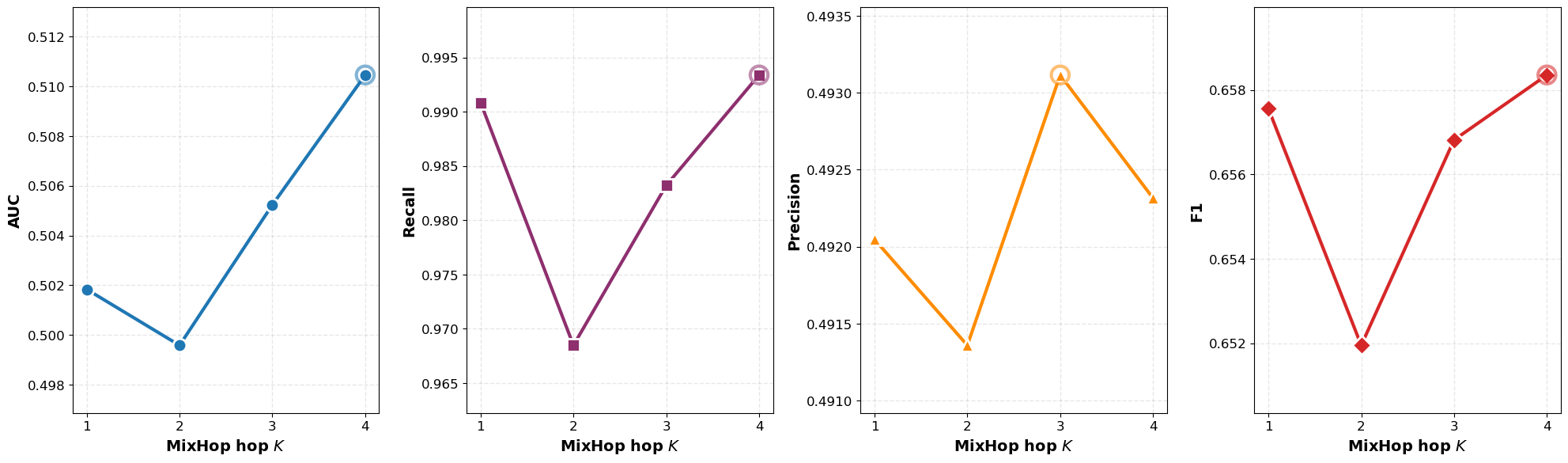}
\caption{Sensitivity analysis with respect to MixHop depth $K$ on the FEMA NFIP dataset.
Performance varies mildly across $K$, indicating that deeper diffusion provides limited
marginal benefit under this domain and that moderate multi-hop aggregation is sufficient.}
\label{fig:fema_K}
\end{figure}

\begin{figure}[h]
\centering
\includegraphics[width=\textwidth]{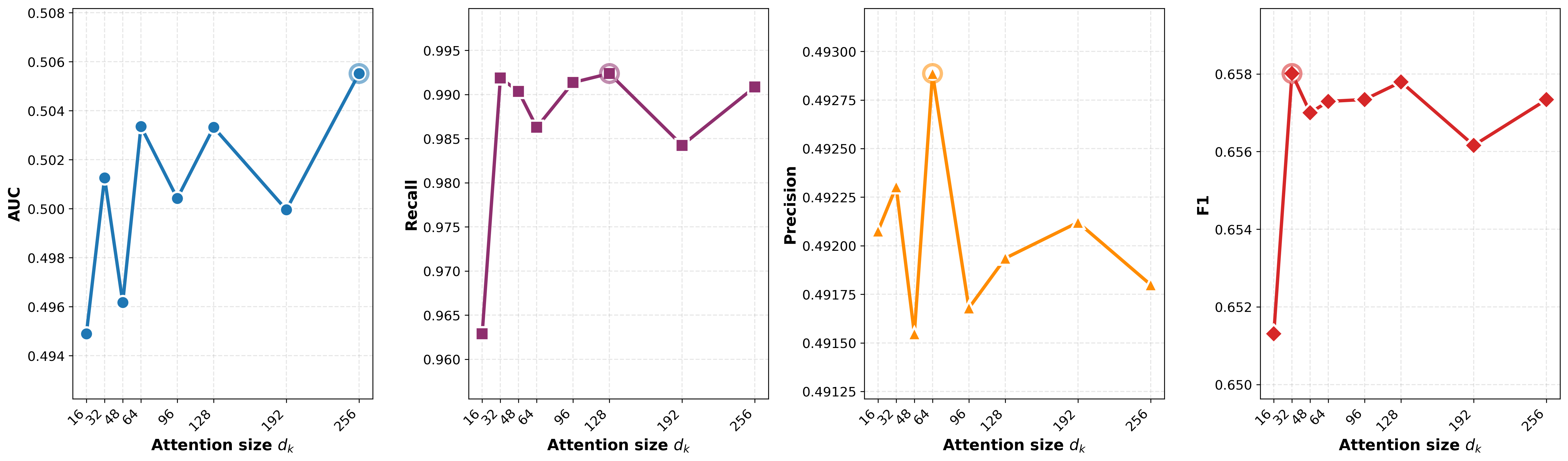}
\caption{Sensitivity analysis with respect to attention dimension $d_k$ on the FEMA NFIP dataset.
Results are relatively stable across a wide range of $d_k$, suggesting diminishing returns
from increasing attention capacity in this setting.}
\label{fig:fema_dk}
\end{figure}

As shown in Figure~\ref{fig:fema_dk},the model exhibits relatively stable behavior across a wide range of temporal attention dimensions. This suggests that temporal coupling in FEMA acts more as a consistency regularizer than as a primary source of discriminative signal, which is reasonable given the weaker short-horizon temporal dynamics in this dataset.

\begin{figure*}[h]
\centering
\includegraphics[width=\textwidth]{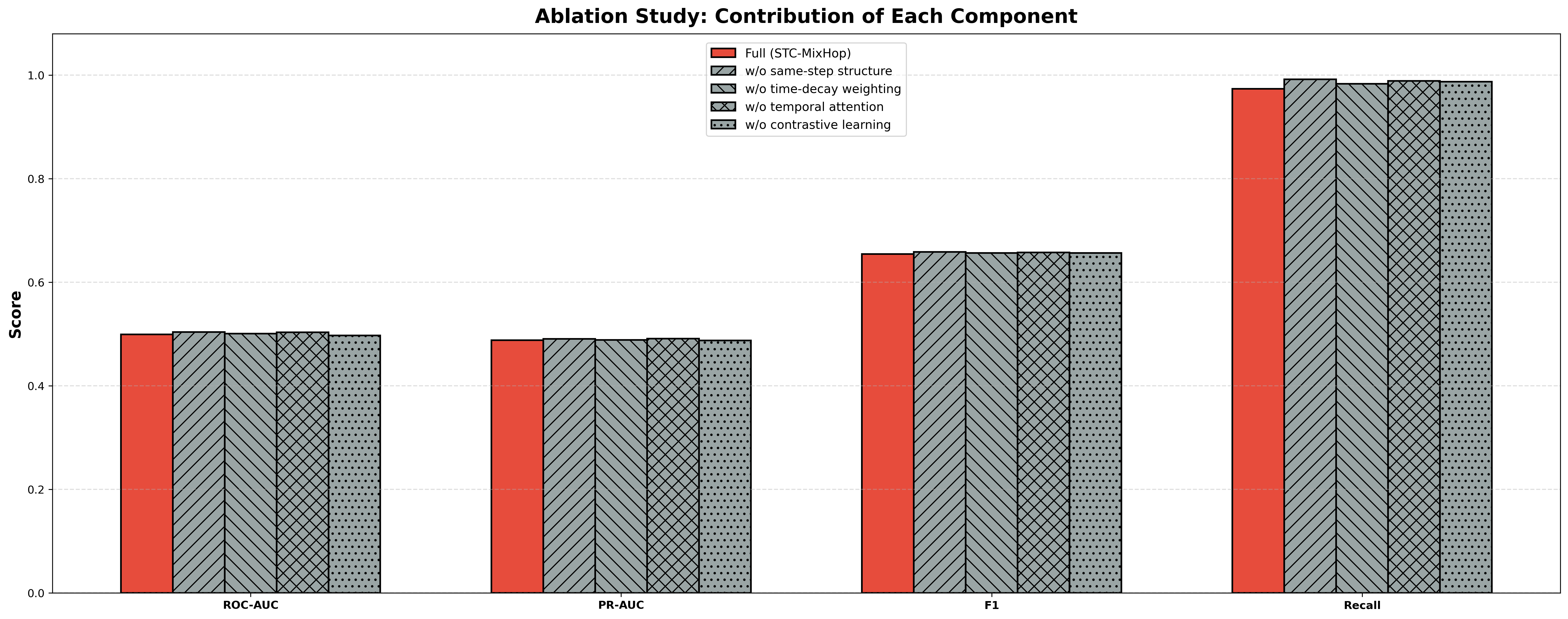}
\caption{Component ablation results on the FEMA NFIP dataset.
Removing individual temporal/contrastive components yields only minor changes,
supporting the view that these modules mainly act as regularizers under this domain.
Interestingly, excluding same-step structural modeling can slightly improve recall,
suggesting potential noise from immediate neighborhood signals in this dataset.}
\label{fig:fema_ablation}
\end{figure*}

The ablation results for the FEMA dataset ( Figure~\ref{fig:fema_ablation} and Table~\ref{tab:fema_ablation}) reveal a comparatively stable performance landscape. Removing individual components changes the metrics only modestly, which suggests that auxiliary modules in this setting mainly contribute regularization and robustness rather than large gains in discriminative power. High recall is preserved across ablations, reinforcing the view that the framework functions primarily as a screening-oriented model on this dataset.
\clearpage

\begin{table}[h!]
\centering
\caption{Ablation study of STC-MixHop components on the FEMA NFIP dataset.
Each variant removes one component while keeping all other settings unchanged.
Differences are relatively small across variants, suggesting that on this dataset
temporal/contrastive components behave primarily as regularizers and that certain
same-step structural signals may introduce noise depending on the domain.}
\label{tab:fema_ablation}
\resizebox{\textwidth}{!}{
\begin{tabular}{lcccccc}
\toprule
Variant & ROC-AUC & PR-AUC & $F_{\beta}$ & F1 & Precision & Recall \\
\midrule
Full (STC-MixHop) & 0.4989 & 0.4879 & 0.5467 & 0.6543 & 0.4927 & 0.9736 \\
w/o same-step structure & 0.5036 & 0.4907 & \textbf{0.5477} & \textbf{0.6582} & 0.4926 & \textbf{0.9919} \\
w/o time-decay weighting & 0.5008 & 0.4886 & 0.5472 & 0.6564 & 0.4926 & 0.9832 \\
w/o temporal attention & 0.5033 & \textbf{0.4911} & 0.5475 & 0.6575 & 0.4925 & 0.9888 \\
w/o contrastive learning & 0.4972 & 0.4876 & 0.5464 & 0.6563 & 0.4915 & 0.9873 \\
\bottomrule
\end{tabular}}
\end{table}

\subsection{Discussion}

The experimental evidence supports several conclusions with implications for both methodology and deployment. First, the behavior of STC-MixHop on PaySim is better understood through the lens of stability, screening utility, and component-level interpretability than through a simple benchmark-style comparison focused only on headline scores. Across ablations, the most reliable contribution comes from multi-hop structural diffusion, while temporal synthesis mainly improves representation stability under chronological evaluation. Contrastive pretraining behaves as a conditional regularizer rather than a universally beneficial performance booster.

Second, the results should be interpreted candidly. On attribute-dominant datasets, strong tabular baselines can remain difficult to outperform. This is particularly visible when informative node or transaction features already separate a large portion of the positive class. In such cases, adding graph structure does not automatically improve performance and may introduce noise if the induced topology is only weakly aligned with the target labels. We view this not as a negative result, but as an important empirical boundary condition that clarifies when graph-based modeling is justified.

Third, the advantage of STC-MixHop is most meaningful in settings where suspicious or anomalous behavior is relationally organized. When risk is associated with hidden topological dependencies, repeated interaction motifs, shared intermediaries, or multi-hop coordination patterns, multi-scale graph propagation provides information that marginal attributes alone cannot fully express. Under these conditions, graph learning contributes most clearly as a high-recall screening tool for difficult and partially concealed cases.

Finally, the study highlights the importance of realistic evaluation design. Strict chronological splits, validation-based threshold selection, and multiple complementary metrics provide a more deployment-relevant assessment than randomized partitioning alone. For fraud and anomaly detection in evolving transaction systems, such protocol choices are not merely implementation details; they materially shape the validity of reported conclusions.

\section{Conclusion}\label{sec:conclusion}

This research presents a comprehensive stability-focused evaluation of the STC-MixHop framework using PaySim, a synthetic mobile money transaction dataset for fraud detection whose data generation properties differ substantially from commonly studied blockchain transaction benchmarks. Our evaluation employs strict chronological segmentation, multiple complementary performance metrics, and systematic ablation studies to provide insights beyond single-dataset optimization.

The experimental results yield three main findings with direct implications for both research methodology and operational deployment. First, multi-resolution topological diffusion is the most important architectural prior within the STC-MixHop framework. Even in attribute-dominant data conditions where extended-range relational patterns are relatively weak, encoding topological context beyond immediate neighbors remains essential for achieving reliable screening-oriented detection behavior. Second, temporal synthesis contributes mainly through representation stabilization rather than introduction of fundamentally new discriminative capacity. By smoothing short-horizon representational drift across temporal snapshots, this component reduces sensitivity to concentrated activity patterns and temporal distributional variation without requiring complex extended-range temporal memory mechanisms. Third, contrastive pretraining serves as a condition-dependent auxiliary regularization component. Although it does not consistently improve ranking-oriented metrics in attribute-dominant conditions, it affects learning dynamics and score distributions in ways that may prove beneficial under different data generation conditions.

The supplementary cross-domain experiments suggest that these component-level findings are not unique to PaySim, although they should be interpreted cautiously because the additional datasets depend more heavily on graph construction choices and were not intended as exhaustive quantitative benchmarks. Across these settings, the most consistent pattern is that structural diffusion remains the strongest contributor, while temporal coupling and auxiliary contrastive objectives mainly affect stability, robustness, and representational smoothness rather than uniformly improving headline metrics.

These findings support a stability-focused perspective on spatial-temporal graph learning for fraud and anomaly detection in evolving transaction systems. Rather than treating architectural components as interchangeable performance boosters, our results show that their functional roles are distinct and condition-dependent. In particular, graph structure is most useful when suspicious behavior is relationally organized or partially concealed, whereas strong tabular baselines remain highly competitive when informative marginal attributes dominate the signal.

\paragraph{Future Directions.}
This research provides a foundation for several promising future directions. While our primary evaluation uses the PaySim corpus, we have also conducted preliminary cross-domain validation on the Porto Seguro and FEMA NFIP datasets (Section~\ref{sec:experiments}). Extending the analysis to additional real-world datasets would further strengthen the empirical basis of the component-level conclusions reported here and help clarify how graph learning behaves under different forms of relational structure, label sparsity, and temporal drift.

The framework's modular architecture also enables straightforward adaptation to domain-specific deployment needs. The attention-based temporal synthesis mechanism provides interpretable weighting coefficients that may support model inspection and documentation in operational settings. In addition, the explicit separation between structural diffusion and classification components makes it easier to incorporate domain constraints, fairness regularization, or customized decision thresholds when required by application context.

From a methodological perspective, adaptive mechanisms that automatically adjust component contributions according to detected data characteristics may improve deployment flexibility. Furthermore, integrating explicit cost-sensitive learning objectives reflecting operational asymmetries between false positives and false negatives would improve the framework's relevance for real-world fraud and anomaly detection settings. We hope this study encourages further work on robust and interpretable graph learning systems for temporally evolving relational risk detection tasks.

\section*{Funding}
This research was funded by the Natural Sciences and Engineering Research Council of Canada (NSERC) under grant number PINV000071561.

\section*{Acknowledgements}
The authors acknowledge the support provided by Gradient Training Funding, including access to decentralized computing infrastructure used in this study.

\section*{Declarations}
\subsection*{Conflict of Interest}
The authors have no competing interests to declare that are relevant to the content of this article.

\subsection*{Ethical Approval}
This article does not contain any studies with human participants or animals performed by any of the authors.

\subsection*{Informed Consent}
Informed consent was not required as this study did not involve human participants.

\subsection*{Data Availability}
The datasets analyzed during the current study are publicly available third-party datasets. The PaySim financial mobile money simulator dataset is available from the Kaggle repository as cited in Reference [1]. The Porto Seguro’s safe driver prediction dataset is available from the Kaggle Competition repository as cited in Reference [36]. The FIMA NFIP redacted policies dataset is available from the OpenFEMA portal as cited in Reference [37]. All persistent web links to these datasets are provided in the references section of this article.

\bibliographystyle{elsarticle-num}
\bibliography{references}

@inproceedings{lopez2016paysim,
  author    = {Lopez-Rojas, Alejandro and Elmir, Bjarne and Axelsson, Stefan},
  title     = {PaySim: A Financial Mobile Money Simulator for Fraud Detection},
  booktitle = {2016 European Modelling Symposium (EMS)},
  year      = {2016},
  publisher = {IEEE},
  note      = {Synthetic mobile money transaction dataset used widely for fraud detection research},
  url       = {https://www.kaggle.com/datasets/ealaxi/paysim1}
}

@techreport{wuthrich2019data,
  author      = {W{\"u}thrich, Mario V. and Buser, Christoph},
  title       = {Data Analytics for Non-Life Insurance Pricing},
  institution = {Swiss Finance Institute},
  year        = {2019},
  number      = {Research Paper No. 16-68},
  note        = {SSRN Manuscript 2870308}
}

@article{vorobyev2024fraud,
  author  = {Vorobyev, Ilya and Kireev, Andrey and Fedulova, Ekaterina},
  title   = {Graph Neural Networks for Financial Fraud Detection: A Survey},
  journal = {ACM Computing Surveys},
  volume  = {56},
  number  = {8},
  pages   = {1--35},
  year    = {2024}
}

@book{frees2014predictive,
  author    = {Frees, Edward W. and Derrig, Richard A. and Meyers, Glenn},
  title     = {Predictive Modeling Applications in Actuarial Science: Volume 1, Predictive Modeling Techniques},
  publisher = {Cambridge University Press},
  year      = {2014},
  series    = {International Series on Actuarial Science}
}

@misc{portoseguro2017,
  author       = {{Porto Seguro}},
  title        = {Porto Seguro's Safe Driver Prediction},
  year         = {2017},
  howpublished = {Kaggle Competition},
  url          = {https://www.kaggle.com/c/porto-seguro-safe-driver-prediction},
  note         = {Brazilian auto insurance claim prediction dataset with 595,212 policies}
}

@misc{fema2024nfip,
  author       = {{Federal Emergency Management Agency}},
  title        = {FIMA NFIP Redacted Policies},
  year         = {2024},
  howpublished = {OpenFEMA Dataset},
  url          = {https://www.fema.gov/openfema-data-page/fima-nfip-redacted-policies-v2},
  note         = {National Flood Insurance Program policy-level data}
}

@article{mohan2023improving,
  author  = {Mohan, A. and PV, K. and Sankar, P. and Maya Manohar, K. and Peter, A.},
  title   = {Improving anti-money laundering in bitcoin using evolving graph convolutions and deep neural decision forest},
  journal = {Data Technologies and Applications},
  volume  = {57},
  number  = {3},
  pages   = {313--329},
  year    = {2023}
}

@article{scarselli2009graph,
  author  = {Scarselli, F. and Gori, M. and Tsoi, A. C. and Hagenbuchner, M. and Monfardini, G.},
  title   = {The graph neural network model},
  journal = {IEEE Transactions on Neural Networks},
  volume  = {20},
  number  = {1},
  pages   = {61--80},
  year    = {2009}
}

@inproceedings{yu2021abnormal,
  author    = {Yu, L. and Zhang, N. and Wen, W.},
  title     = {Abnormal transaction detection based on graph networks},
  booktitle = {2021 IEEE 45th Annual Computers, Software, and Applications Conference (COMPSAC)},
  pages     = {312--317},
  year      = {2021},
  publisher = {IEEE}
}

@inproceedings{abuelhaija2019mixhop,
  author    = {Abu-El-Haija, S. and Perozzi, B. and Kapoor, A. and Harutyunyan, H. and Alipourfard, N. and Lerman, K. and Steeg, G. V. and Galstyan, A.},
  title     = {Mixhop: Higher-order graph convolutional architectures via sparsified neighborhood mixing},
  booktitle = {Proceedings of the 36th International Conference on Machine Learning (ICML)},
  pages     = {21--29},
  year      = {2019},
  publisher = {PMLR},
  url       = {http://proceedings.mlr.press/v97/abu-el-haija19a.html}
}

@article{wang2025cosemignn,
  author  = {Wang, Yulong and Zheng, Qingxiao and Li, Xuedong and Wang, Lingfeng and Lin, Ling},
  title   = {{CoSemiGNN}: Blockchain fraud detection with dynamic graph neural networks based on co-association of semi-supervised},
  journal = {Expert Systems with Applications},
  volume  = {298},
  pages   = {129853},
  year    = {2025},
  doi     = {10.1016/j.eswa.2025.129853}
}

@article{chen2025mdst,
  author  = {Chen, S. and Liu, Y. and Zhang, Q. and Shao, Z. and Wang, Z.},
  title   = {Multi-distance spatial-temporal graph neural network for anomaly detection in blockchain transactions},
  journal = {Advanced Intelligence System},
  volume  = {7},
  number  = {8},
  year    = {2025}
}

@article{lo2023inspection,
  author  = {Lo, Wai Weng and Kulatilleke, Gayan K. and Sarhan, Mohanad and Layeghy, Siamak and Portmann, Marius},
  title   = {Inspection-L: Self-supervised GNN node embeddings for money laundering detection in bitcoin},
  journal = {Applied Intelligence},
  volume  = {53},
  pages   = {19406--19417},
  year    = {2023},
  doi     = {10.1007/s10489-023-04504-9}
}

@article{li2021selfsupervised,
  author  = {Li, S. and Xu, F. and Wang, R. and Zhong, S.},
  title   = {Self-supervised incremental deep graph learning for ethereum phishing scam detection},
  year    = {2021}
}

@inproceedings{weber2019gcn,
  author    = {Weber, Mark and Domeniconi, Giacomo and Chen, Jie and Weidele, Daniel Karl I. and Bellei, Claudio and Robinson, Tom and Leiserson, Charles E.},
  title     = {Anti-money laundering in bitcoin: Experimenting with graph convolutional networks for financial forensics},
  booktitle = {KDD '19 Workshop on Anomaly Detection in Finance},
  year      = {2019},
  url       = {https://arxiv.org/abs/1908.02591}
}

@inproceedings{pareja2020evolvegcn,
  author    = {Pareja, A. and Domeniconi, G. and Chen, J. and Ma, T. and Suzumura, T. and Kanezashi, H. and Kaler, T. and Schardl, T. B. and Leiserson, C. E.},
  title     = {EvolveGCN: Evolving graph convolutional networks for dynamic graphs},
  booktitle = {Proceedings of the Thirty-Fourth AAAI Conference on Artificial Intelligence},
  year      = {2020}
}

@inproceedings{dai2024tgat,
  author    = {Dai, C. and Tang, Q. and Ding, H.},
  title     = {TGAT: Temporal graph attention network for blockchain phishing scams detection},
  booktitle = {2024 International Conference on Computer, Information and Telecommunication Systems (CITS)},
  pages     = {1--7},
  year      = {2024},
  publisher = {IEEE},
  address   = {Girona, Spain},
  doi       = {10.1109/CITS61189.2024.10608015}
}

@inproceedings{velickovic2019deep,
  author    = {Veli{\v{c}}kovi{\'c}, Petar and Fedus, William and Hamilton, William L. and Li{\`o}, Pietro and Bengio, Yoshua and Hjelm, R. Devon},
  title     = {Deep Graph Infomax},
  booktitle = {International Conference on Learning Representations (ICLR)},
  year      = {2019},
  url       = {https://openreview.net/forum?id=rklz9iAcKQ}
}

@inproceedings{velickovic2018gat,
  author    = {Veli{\v{c}}kovi{\'c}, Petar and Cucurull, Guillem and Casanova, Arantxa and Romero, Adriana and Lio, Pietro and Bengio, Yoshua},
  title     = {Graph Attention Networks},
  booktitle = {International Conference on Learning Representations (ICLR)},
  year      = {2018},
  url       = {https://openreview.net/forum?id=rJXMpikCZ}
}

@inproceedings{you2020graph,
  author    = {You, Yuning and Chen, Tianlong and Sui, Yongduo and Chen, Ting and Wang, Zhangyang and Shen, Yang},
  title     = {Graph contrastive learning with augmentations},
  booktitle = {Advances in Neural Information Processing Systems (NeurIPS)},
  volume    = {33},
  pages     = {5812--5823},
  year      = {2020},
  url       = {https://proceedings.neurips.cc/paper/2020/file/3fe230348e9a12c13120749e3f9fa4cd-Paper.pdf}
}

@article{reynisson2024graphguard,
  author  = {Reynisson, K. and Schreyer, M. and Borth, D.},
  title   = {GraphGuard: Contrastive Self-Supervised Learning for Credit-Card Fraud Detection in Multi-Relational Dynamic Graphs},
  journal = {arXiv preprint arXiv:2407.12440},
  year    = {2024},
  url     = {https://arxiv.org/abs/2407.12440}
}

@inproceedings{hamilton2017graphsage,
  author    = {Hamilton, William L. and Ying, Rex and Leskovec, Jure},
  title     = {Inductive Representation Learning on Large Graphs},
  booktitle = {Advances in Neural Information Processing Systems (NeurIPS)},
  volume    = {30},
  year      = {2017},
  url       = {https://proceedings.neurips.cc/paper/2017/hash/5dd9db5e033da9c6fb5ba83c7a7ebea9-Abstract.html}
}

@inproceedings{sun2024mcre,
  author    = {Sun, Tengjiao and Li, Xiang and Shi, Tianyu and Peng, Jiahui and Zheng, Sheng and Kim, Hansung},
  title     = {MCRE: Multimodal Conditional Representation and Editing for Text-Motion Generation},
  booktitle = {European Conference on Computer Vision (ECCV)},
  pages     = {406--414},
  year      = {2024},
  organization = {Springer}
}

@article{sun2025objective,
  author  = {Sun, Wenxi and Shen, Qiannan and Gao, Yijun and Mao, Qinkai and Qi, Tongsong and Xu, Shuo},
  title   = {Objective over Architecture: Fraud Detection Under Extreme Imbalance in Bank Account Opening},
  journal = {Computation},
  year    = {2025},
  volume  = {13},
  number  = {12},
  pages   = {290},
  doi     = {10.3390/computation13120290},
  url     = {https://www.mdpi.com/2079-3197/13/12/290}
}

@misc{shensunqi2025,
  author       = {Sun, Wenxi and Qi, Zhichun and Shen, Qiannan},
  title        = {High-Recall Deep Learning: A Gated Recurrent Unit Approach to Bank Account Fraud Detection on Imbalanced Data},
  year         = {2025},
  howpublished = {Research Square Preprint},
  doi          = {10.21203/rs.3.rs-8136120},
  url          = {https://www.researchsquare.com/article/rs-8136120}
}

@inproceedings{ke2025detection,
  author    = {Ke, Zong and Zhou, Shicheng and Zhou, Yining and Chang, Chia Hong and Zhang, Rong},
  title     = {Detection of AI Deepfake and Fraud in Online Payments Using GAN-Based Models},
  booktitle = {2025 8th International Conference on Advanced Algorithms and Control Engineering (ICAACE)},
  pages     = {1786--1790},
  year      = {2025},
  organization = {IEEE},
  doi       = {10.1109/ICAACE65325.2025.11020513}
}

@inproceedings{yu2024identifying,
  author    = {Yu, Qian and Ke, Zong and Xiong, Guofu and Cheng, Yu and Guo, Xiaojun},
  title     = {Identifying Money Laundering Risks in Digital Asset Transactions Based on {AI} Algorithms},
  booktitle = {2024 4th International Conference on Electronic Information Engineering and Computer Communication (EIECC)},
  pages     = {1081--1085},
  year      = {2024},
  month     = {December},
  organization = {IEEE},
  url       = {https://ieeexplore.ieee.org/document/10929087}
}
\end{document}